%% file: example.tex
\newcommand{\msssim}{\text{msssim}}

\newcommand{\Tab}{T_{A\rightarrow B}}
\newcommand{\Tba}{T_{B\rightarrow A}}
\newcommand{\pih}{\hat{\pi}}
\newcommand{\sh}{\hat{s}}
\newcommand{\ph}{\hat{p}}
\newcommand{\pis}{\pi,s} %
\newcommand{\pish}{\pih,\sh} %
\newcommand{\ppis}{(\pis)} %
\newcommand{\ppish}{(\pish)} %

\newcommand{\toN}{^{1\dots N}}

\newcommand{\toL}{^{1\dots L}}

\newcommand{\lUC}{\ell_{UC}}
\newcommand{\lTR}{\ell_{TR}}
\newcommand{\eqand}{\ \ \ \text{ and }\ \ \ }

\newcommand{\real}{\mathbb{R}}

\newcommand{\similar}{\Omega}
\newcommand{\unprojection}{\omega}
\newcommand{\expectation}{\mathop{\mathbb{E}}}
\newcommand{\lsum}{\sum\limits}

\documentclass{article}

\usepackage[final]{corl_2022} %
\usepackage{graphicx}
\usepackage{float}
\usepackage{amsfonts}
\usepackage{amsmath}
\usepackage{multicol}
\usepackage{multirow}
\usepackage[utf8]{inputenc}
\usepackage[T1]{fontenc}
\usepackage{array, booktabs, ragged2e}
\usepackage{comment}
\usepackage{siunitx}
\usepackage{hyperref}

\title{Neural Neural Textures Make Sim2Real Consistent}

\author{%
	Ryan Burgert \quad Jinghuan Shang \quad Xiang Li \quad Michael S. Ryoo\\
	Department of Computer Science\\
	Stony Brook University\\
	Stony Brook, NY 11794 \\
	\texttt{\{rburgert,jishang,xiangli8,mryoo\}@cs.stonybrook.edu} \\
}

\begin{document}
\maketitle

\vspace{-20pt}
\begin{abstract}
	Unpaired image translation algorithms can be used for sim2real tasks, but many fail to generate temporally consistent results.
	We present a new approach that combines differentiable rendering with image translation to achieve temporal consistency over indefinite timescales, using surface consistency losses and \emph{neural neural textures}.
	We call this algorithm TRITON (Texture Recovering Image Translation Network): an unsupervised, end-to-end, stateless sim2real algorithm that 
	leverages the underlying 3D geometry of input scenes by generating realistic-looking learnable neural textures.
	By settling on a particular texture for the objects in a scene, we ensure consistency between frames statelessly.
	TRITON is not limited to camera movements --- it can handle the movement and deformation of objects as well, making it useful for downstream tasks such as robotic manipulation.
		We demonstrate the superiority of our approach both qualitatively and quantitatively, using robotic experiments and comparisons to ground truth photographs. We show that TRITON generates more useful images than other algorithms do. Please see our project website: \href{https://tritonpaper.github.io}{tritonpaper.github.io}
\end{abstract}
\keywords{sim2real, image translation, differentiable rendering} 

\section{Introduction}

Current sim2real image translation algorithms used for robotics \cite{retina_gan,rl_cyclegan} often have difficulty generating consistent results across large time-frames particularly when the objects in an environment are allowed to move. This makes training good robotic policies using such sim2real challenging.
In this paper, we discuss an algorithm called TRITON (Texture Recovering Image Translation Network) that combines neural rendering, image translation, and two special surface consistency losses to create surface-consistent translations over frames.
We use the term surface-consistent to refer to the desirable quality of preserving the visual appearance of object surfaces as they move, or are viewed from another angles. 

TRITON is applicable when we have a simulator capable of generating a realistic distribution of geometric data, but when we do not have any information about the surfaces of those objects (which we need to render realistic images). For example, in one of our experiments we use a robotic arm model provided by the manufacturer that contains no material information, and then learn to render the materials of the arm using TRITON.
As opposed to requiring an expert/artist to create 3D models with meaningful textures, TRITON can generate these from scratch using a set of photographs capturing the distribution of states in a simulation, without requiring any matches between domains. 
That is, given a set of unpaired, unannotated real images and geometry images, TRITON learns the underlying texture of objects and surfaces appearing in the scene without any direct supervision.

We introduce the concept of \emph{neural neural texture}, which is an implicit texture representation having the form of a neural network function generating RGB values given surface coordinates.
Unlike previous raster-based `neural textures' to create realistic images \cite{deferred_neural_rendering,surgical_video_translation}, we model surface textures as a function instead of discretized pixels.
It is a continuous implicit pixel-less parametric texture represented by a neural network using Fourier features \cite{fourier_feature_networks} that takes in 2D UV coordinates and outputs colors. This is similar to how neural radiance fields (NeRF) represents a 3D scene in the form of a neural function \cite{nerf}. The difference is that we focus on learning textures from unpaired training images of very different scene configurations, for viewpoint as well as object motion synthesis.

We conduct multiple experiments to confirm TRITON's advantage over prior image translation approaches commonly used for sim2real including CycleGAN~\cite{cyclegan} and CUT~\cite{cut}.
Importantly, we show the advantages of the proposed approach in real-robot sim2real experiments, learning the textures and training a robot policy solely based on the images generated by TRITON. 
\\
\begin{figure}[t]
	\vspace{-10pt}
	\begin{center}
		\includegraphics[width=0.99\textwidth]{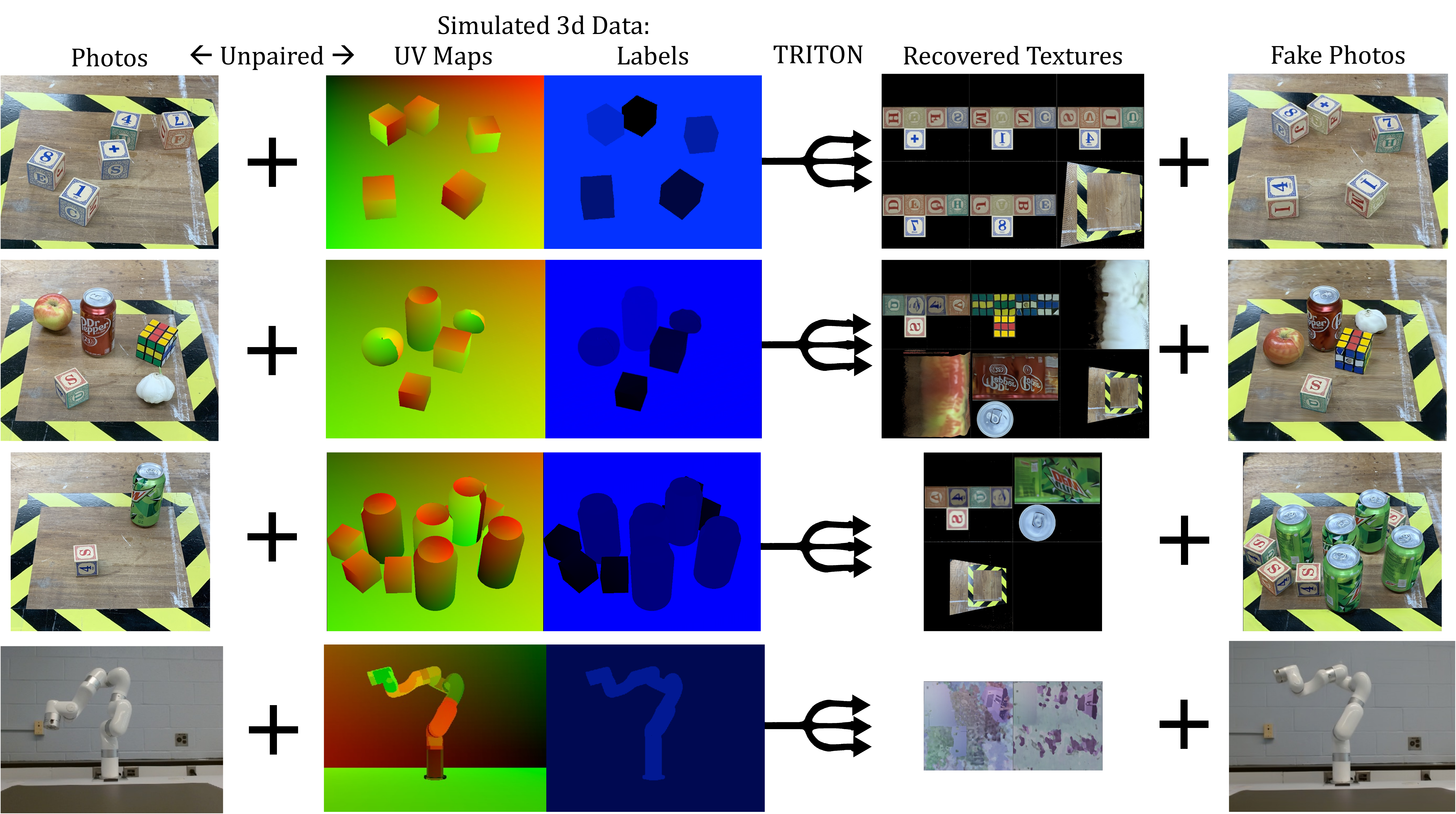}
	\end{center}
	\vspace{-5pt}
	\caption{
		This is an overview of what TRITON accomplishes.
		TRITON learns the textures of 3D objects to help translate simulated images into photographs. It does this with unpaired data. Each row is a different dataset.
	}
	\label{fig:first_diagram}
	\vspace{-5pt}
\end{figure}
	
\vspace{-15pt}	
\section{Related Works}
\vspace{-3pt}
\paragraph{Unsupervised Image to Image and Video to Video Translation}
Generative models have been very successful in creating images unconditionally~\cite{gan,stylegan}, creating images conditioned on text~\cite{dalle,imagen} and creating images when trained on paired data~\cite{pixtopix}.
Unpaired image translation algorithms include ~\cite{dual_diffusion,cyclegan,munit,unit}.
Of particular interest are sim2real image to image translation algorithms such as~\cite{retina_gan,rl_cyclegan,surgical_image_translation,enhancing_photorealism_enhancement}. These algorithms aim to translate simulated images into realistic ones, for various purposes such as data augmentation or video game enhancement. In particular, \citet{retina_gan,rl_cyclegan} are used to help train robots. Our paper uses an architecture based on~\citet{surgical_image_translation}, which is based on MUNIT~\cite{munit}.
Extending to video to video translation, optical flow is often used for stateful translation to improve temporal coherence ~\cite{tecogan,mocyclegan}. But they do not provide temporal coherence over long time periods.%

\vspace{-8pt}
\paragraph{View Synthesis}
NeRF~\cite{nerf} has spawned a large body of research on end-to-end view synthesis like \cite{nerv,fast_nerf,barf}.
Related to our work, which also uses geometry based backbone are \cite{free_view_synthesis,stable_view_synthesis,ners}. However, all of these works only work with static scenes. There are NeRF-based approaches that work on dynamic scenes~\cite{dnerf,nerfies,nerualradianceflow} but none of these were interactive, prohibiting their usage for sim2real. \citet{playable_environments} is interactive, but its conditioned on learned actions and not compatible with physics simulators that would be used to train robots.
To learn the dynamics in the context of robot learning, self-supervised learning over time is an available approach~\cite{tcn,disentangle,3dtrl,nerftcn}. Differently, we only use the geometry without temporally aligned samples in those works.

\vspace{-8pt}
\paragraph{Neural Textures}
\citet{deferred_neural_rendering} introduced the concept of neural textures and image translators in the context of a deferred rendering pipeline, used in other works such as \cite{surgical_video_translation} which is closely related to our project. 
\citet{surgical_video_translation} differs from TRITON in a few very important aspects though, the biggest being that it only synthesizes new camera views, where the only difference between each scene is the placement of the camera. In contrast, our algorithm takes in scenes where several objects are placed in different places or deformed, where one static global 3D model of the environment is not sufficient to accomplish our tasks. In addition, TRITON introduces a new type of neural texture, ``neural neural texture'', which is represented implicitly using a Fourier feature network \cite{fourier_feature_networks}.

\section{Formulation}
\label{sec:data}
	TRITON's goal is to turn 3D simulated images into realistic fake photographs (by training it without any matching pairs), while maintaining high surface consistency.
	It does this by simultaneously learning both an image translator and a set of realistic textures.
	These translations can be useful for downstream tasks, especially robotic policy learning from camera data, enabling sim2real.
	
	In contrast to previous works involving neural textures \cite{deferred_neural_rendering,surgical_video_translation}, TRITON can handle more than just camera movements:
	the positions of objects in the training data can be moved around or deformed as well between data samples.
	With high surface consistency, surfaces of translated objects will look the same even when moved around or viewed from different camera angles.

	\begin{figure}[t]
	    \vspace{-5pt}
		\begin{center}
			\includegraphics[width=\textwidth]{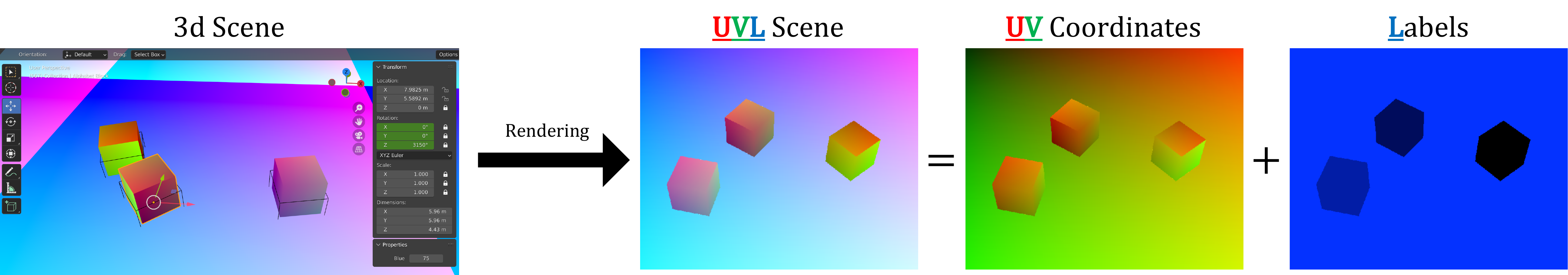}
		\end{center}
		\vspace{-10pt}
		\caption{
			A scene is obtained by rendering a 3D state into an image, where the first two color channels represent $u,v$ coordinates and the third channel $l$ represents object labels.
			}
		\label{fig:uvl_explanation}
		\vspace{-5pt}
	\end{figure}

	There are two components in every dataset: a set of simulated 3D scenes and a set of photographs.
	Datasets usually have many more scenes than photographs, since scenes can be generated automatically.
	A photograph is an image tensor, having $(r,g,b)$ values between 0 and 1. We denote it 
	  as ${p \in [0,1]^{H \times W \times 3}}$ where 
		3 refers to the three $(r,g,b)$ channels, 
		and $H,W$ are the height and width.
		
	In this work, a scene refers to a rendered 3D simulation state - which includes the position, rotation, and deformation of every object (including the camera).
	We represent each 3D state with a simple $(r,g,b)$ image which we call a UVL scene, which is simple enough format that any halfway decent simulator should able to provide.
	A scene has the same dimensions as a photograph and is denoted as 
	$s \in [0,1]^{H \times W \times 3}$.
	However, unlike $p$, $s$'s three $(r,g,b)$ channels encode a special semantic meaning: $(u,v,l)$ as depicted in Figure \ref{fig:uvl_explanation}.
	The channels $u$ and $v$ refer to a texture coordinate, whereas $l$ refers to a label value that identifies every object in a scene.
	In a given dataset, each object gets a different label. Likewise, each object is assigned a different texture.
	We assume that every dataset has $L$ different label values for $L$ different objects, and that we must learn $L$ different neural textures.

        \begin{figure}[bth]
		\begin{center}
			\includegraphics[width=\textwidth]{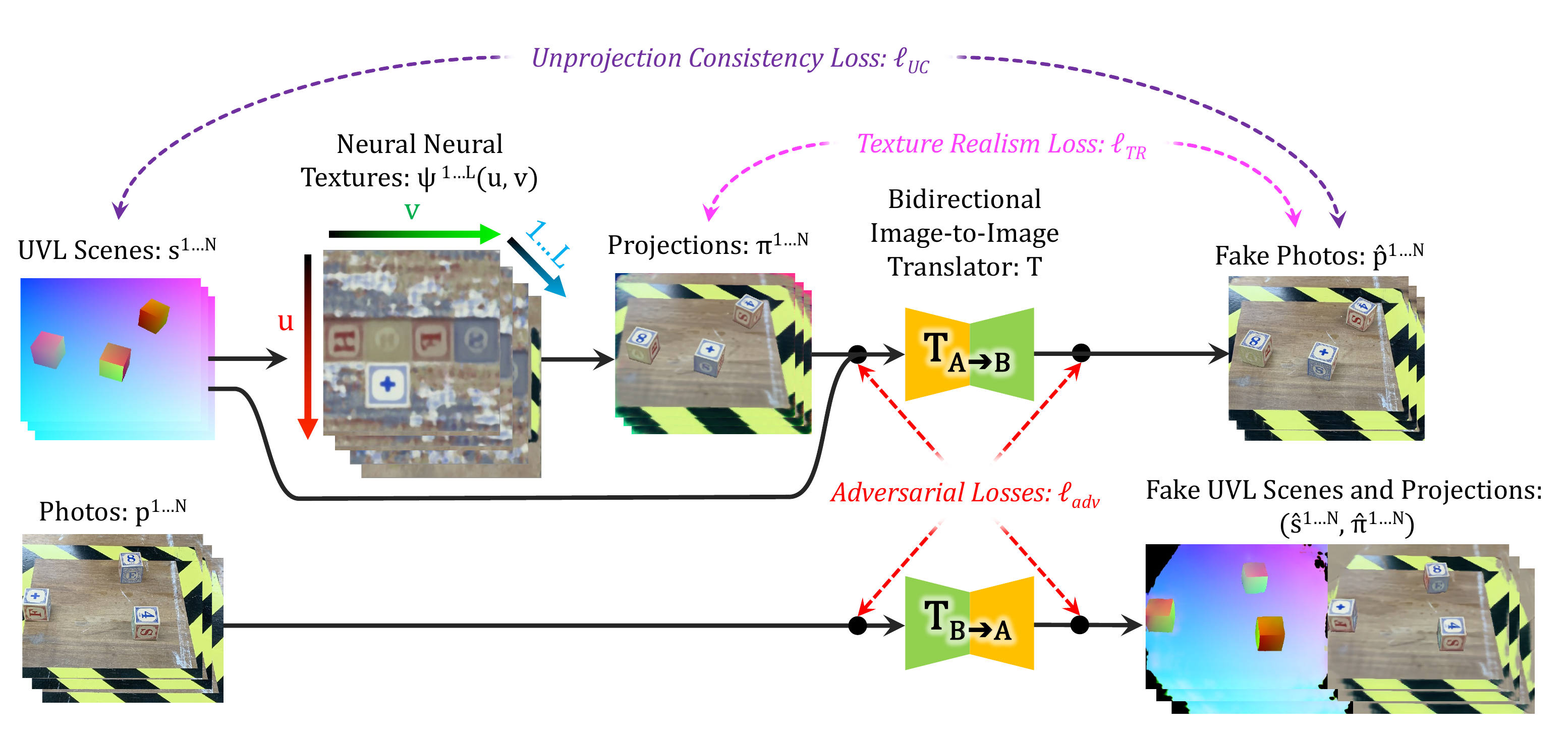}
		\end{center}
		\caption{
			An overview of TRITON.
			Inputs are on the left, and outputs are on the right.
			The dashed arrows indicate losses.
			Although $\pi\toN$ and $\ph\toN$ look similar, they are not identical - $\lTR$ encourages them to look as similar as possible.
		}
		\label{fig:main_diagram}
	\end{figure}

\section{Method}\label{sec:method}
		In this section, we describe how TRITON works to translate images from domain A (simulation) to domain B (photos) while maintaining surface consistency (Figure \ref{fig:main_diagram}). 
		TRITON introduces a learnable neural neural texture with two novel surface consistency losses to an existing image translator. In the end, TRITON is able to effectively generate photorealistic images.
 
\vspace{-3pt}
	\subsection{Neural Neural Texture Projection}
	\label{sec:neural_tex}
	\vspace{-3pt}

		Instead of feeding a UVL scene $s$ directly into the image translator $\Tab$, we first apply learnable textures $\psi$ to the object surfaces, which is learned jointly with $\Tab$. 
		
		These neural neural textures are represented implicitly by a neural network that maps UV values to RGB values. %
		For each texture $\psi^i \in \psi\toL$, 
		\begin{equation}
			\psi^i: (u,v) \in\real^2 \rightarrow (r,g,b) \in\real^3 %
		\end{equation}
		Given a UVL scene $s$, we obtain projection $\pi$ by applying a texture $\psi^i \in \psi\toL$ to every pixel $(u,v,l) \in s$ individually, where texture index $i=I(l)$ is decided by the label value $l$ of that scene pixel. The projection $\pi$, which is an intermediate image is computed as:
		\begin{equation}
			\pi_{[x,y]}=\psi^i(s_{[x,y,1]},s_{[x,y,2]})
		\end{equation}
		where $ x\in \{1\dots W\},\ y\in \{1\dots H\}$. Texture index $i = I(l)$ where $l=s_{[x,y,3]}$ and the function $I(\cdot)$ scales and discretizes $l$ into integers. Subscripts mean multidimensional indexing.

		\begin{figure}[t]
		    \vspace{-5pt}
			\begin{center}
				\includegraphics[width=0.7\textwidth]{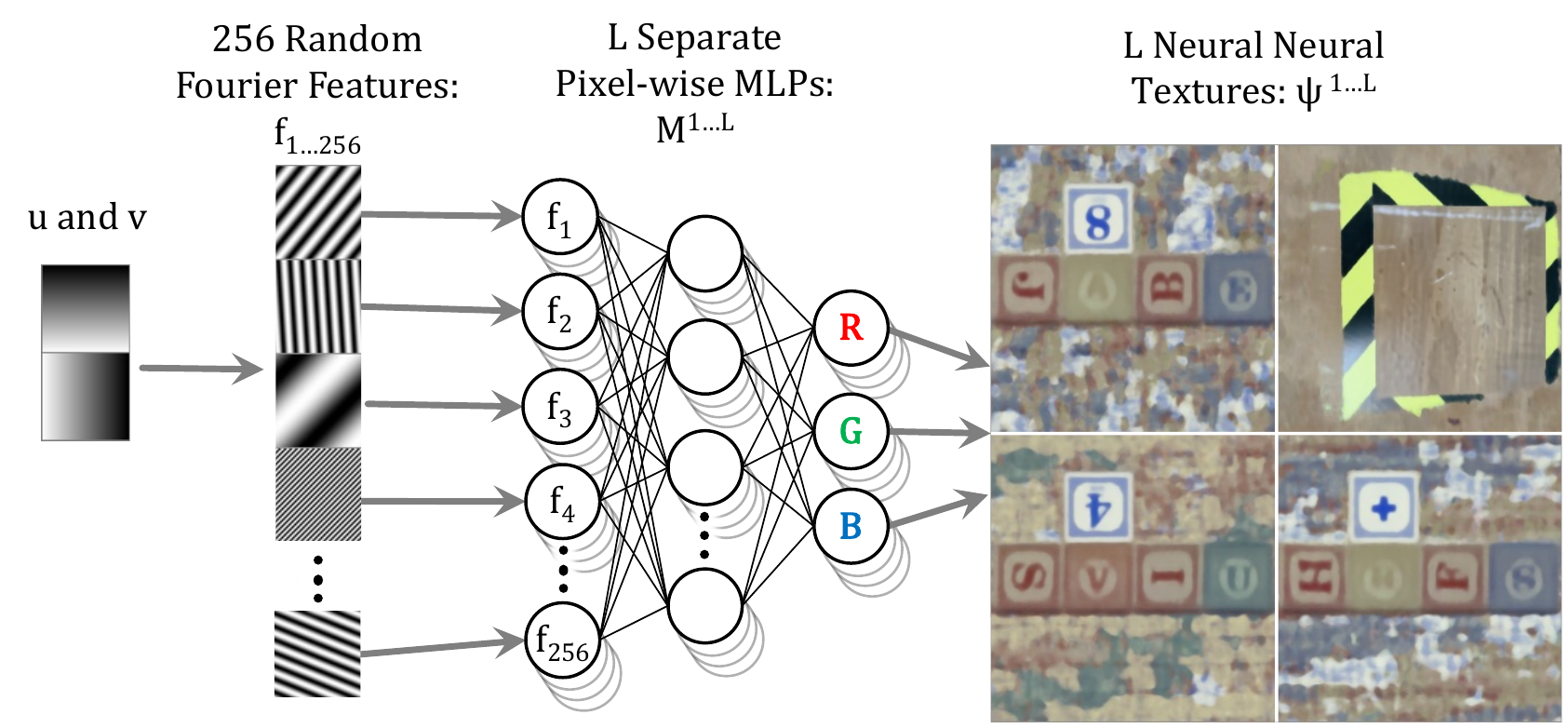}
			\end{center}
			\vspace{-3pt}
			\caption{
				Details of calculating neural neural textures in Figure~\ref{fig:main_diagram}. They are fully differentiable, and represented continuously along the texture's u and v axes using Fourier feature networks.
			}
			\vspace{-8pt}
			\label{fig:learnable_textures}
		\end{figure}

		We now discuss the implementation of our neural neural networks $\psi\toL$.
		Each neural neural texture $\psi^i$ is a function consisting of two components: a multi-layer perceptron $M^i$ and a static set of 256 random spatial frequencies $f_{1...256} \in \real ^ {256 \times 2}$ (Figure \ref{fig:learnable_textures}). 
		Given a two-dimensional $(u,v)$ vector, the spatial frequencies are used to generate a high dimensional embedding vector of $(u,v)$: $\left(\sin(g_1), \cos(g_1), \ldots, \sin(g_{256}), \cos(g_{256})\right)$     . These embeddings are fed into $M^i$, where $M^i$ is a function $\real^{512} \to \real^{3}$ mapping that embedding to an $(r,g,b)$ color value:
		\begin{equation}
				\psi^i(u,v) = 
				\
				M^i(
					\sin( g_1      ) , \ 
					\cos( g_1      ) , \ 
					\sin( g_2      ) , \ 
					\cos( g_2      ) , \ 
					\dots \ 
					\sin( g_{256}  ) , \ 
					\cos( g_{256}  )  \ 
				)
		\end{equation}
		where $g_k=f_k\cdot(u,v)$.
		$f_k = (\alpha_k, \beta_k)$ is a two dimensional vector, where $\alpha_k, \beta_k \in \mathcal{N}(\mu=0,\ \sigma=10)$
		\mbox{and $\cdot$} is the dot product operator.
		The hyperparamters 256 and 10 are selected empirically.

		In practice, we found that TRITON learns faster and more stably with our neural neural textures than it does with raster neural textures. The generated textures are also smoother and less noisy. See the supplementary material for more details.

\vspace{-3pt}
	\subsection{Image Translation}
	\label{sec:image_translation}
\vspace{-3pt}

		Our image translation module $T$ is bidirectional: $\Tab$ translates sim to real (aka domain A to domain B), and $\Tba$ translates real to sim (aka domain B to domain A).
		\begin{equation}
			\ph=\Tab(\pis)  \eqand  \ppish=\Tba(p)
		\end{equation}
			
		$T$ is based on the image translation module used in \citet{surgical_video_translation}, which is a based on a variant \cite{surgical_image_translation} of MUNIT \cite{munit}. 
		$T$ uses the same network architectures as MUNIT, and also inherits its adversarial losses $\ell_{adv}$, cycle consistency losses $\ell_{cyc}$ and reconstruction losses $\ell_{rec}$.
		The main differences between our image translation model and MUNIT are that we use a different image similarity loss $\Omega$. Like \citet[]{surgical_image_translation} and \citet[]{surgical_video_translation} our style code is fixed (making it unimodal) and noise is injected into the latent codes during training to prevent overfitting. 
		During inference however, this intermediate noise is removed and our image translation module is deterministic. 

		MUNIT translates images by encoding both domains $A$ and $B$ into a common latent space, then decoding them into their respective domains  $B$ and $A$.
		In our translation module (Figure \ref{fig:main_diagram}), we define fake photos $\ph = \Tab(\pis) = G_B(E_A(\pis))$ and fake projection/UVL scenes $\ppish = \Tba(p) = G_A(E_B(p))$, where $E_A, E_B, G_A, G_B$ are encoders and generators for domains A and B respectively.

		We define an image similarity loss $\similar$ between two images $x,y$ that returns a score between -1 and 1, where 0 means perfect similarity:
		\begin{equation}
		\label{eq:omega}
			\similar(x,y) = L_2(x,y)-\msssim(x,y) 
		\end{equation}
		This function combines mean pixel-wise $L_2$ distance (used in MUNIT) with multi-scale structural image similarity $\msssim$, introduced in \citep{msssim}. Note that this loss can also be applied to the latent representations obtained by $E_A$ and $E_B$, because like images those tensors are also three dimensional.

		We have cycle consistency loss ${\ell_{cyc}  =  \similar\left(\ppis,T_{B\rightarrow A}\left(\hat{p}\right)\right)  +  \similar\left(p,T_{A\rightarrow B}\left(\hat{\pi},\hat{s}\right)\right)}$
		, similarity loss ${\ell_{rec} = \similar\left(\ppis,G_A\left(E_A\ppis\right)\right) + \similar\left(p,G_B\left(E_B\left(p\right)\right)\right)}$
			(which is effectively an autoencoder loss),   
		and content similarity loss ${ \ell_{con} = \similar\left(E_A\left(\pis\right), E_B\left(\ph\right) \right)    +   \similar\left(E_A\left(p\right),E_A\left(\pish\right)\right) }$.
		We also have adversarial losses $\ell_{adv}$ that come from two discriminators $D_A$ and $D_B$, targeting domains A and B respectively, using the LS-GAN loss introduced by \citet{lsgan}. 
		In total, our image translator loss is $\ell_T=\ell_{cyc}+\ell_{rec}+\ell_{con}+\ell_{adv}$.

\begin{figure}
    \centering
    \begin{minipage}{0.57\textwidth}
		\includegraphics[width=1\linewidth]{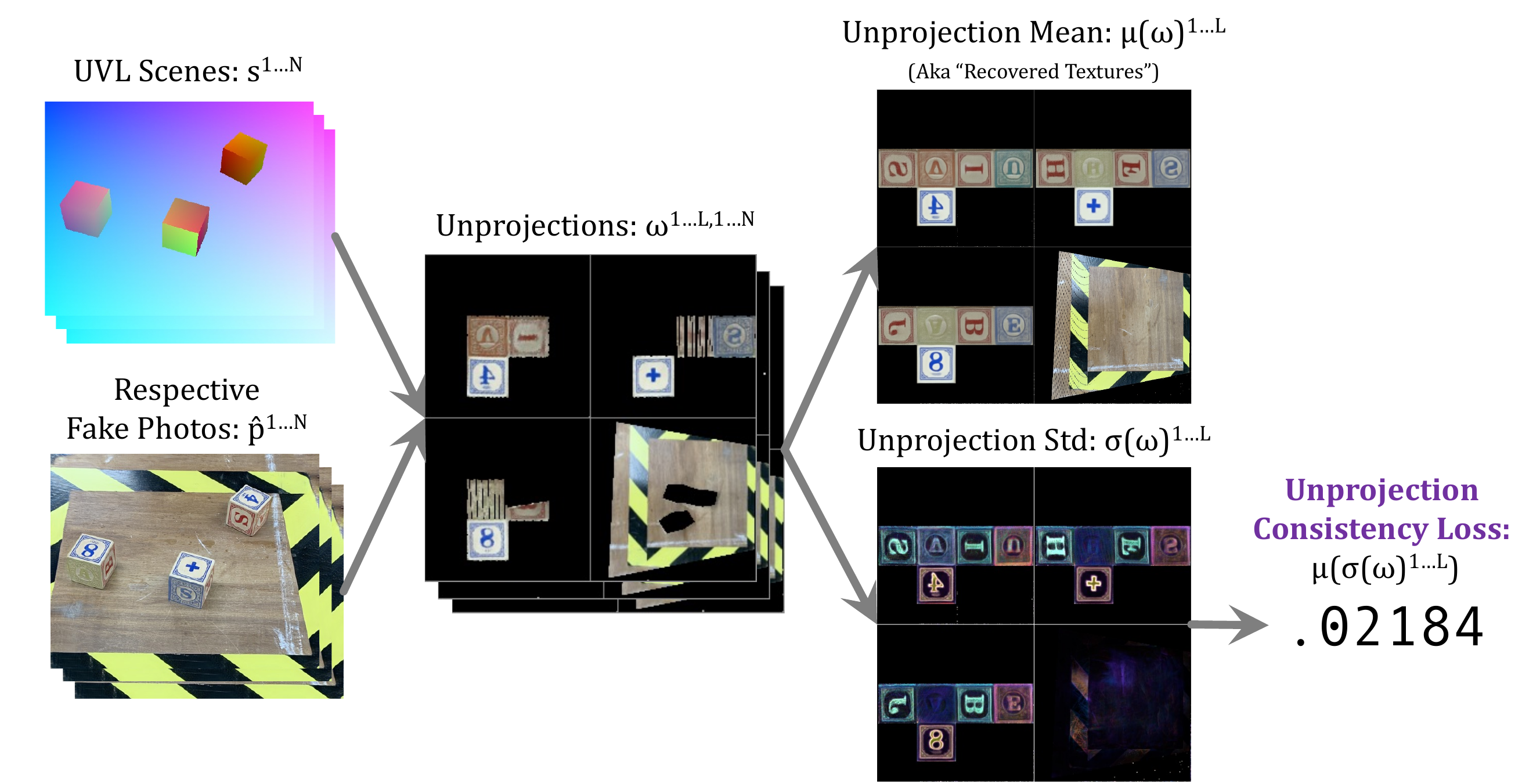}
		\vspace{-5pt}
	\caption{
		This is the unprojection consistency loss used in Figure~\ref{fig:main_diagram}.
		The Unprojection Mean is not used in any losses, but help to illustrate the content in Section \ref{sec:texture_realism_loss}.
	}
	\label{fig:unprojection_consistency_loss}
	\end{minipage}
	\hfill
    \begin{minipage}{0.39\textwidth}
        \includegraphics[width=1\linewidth]{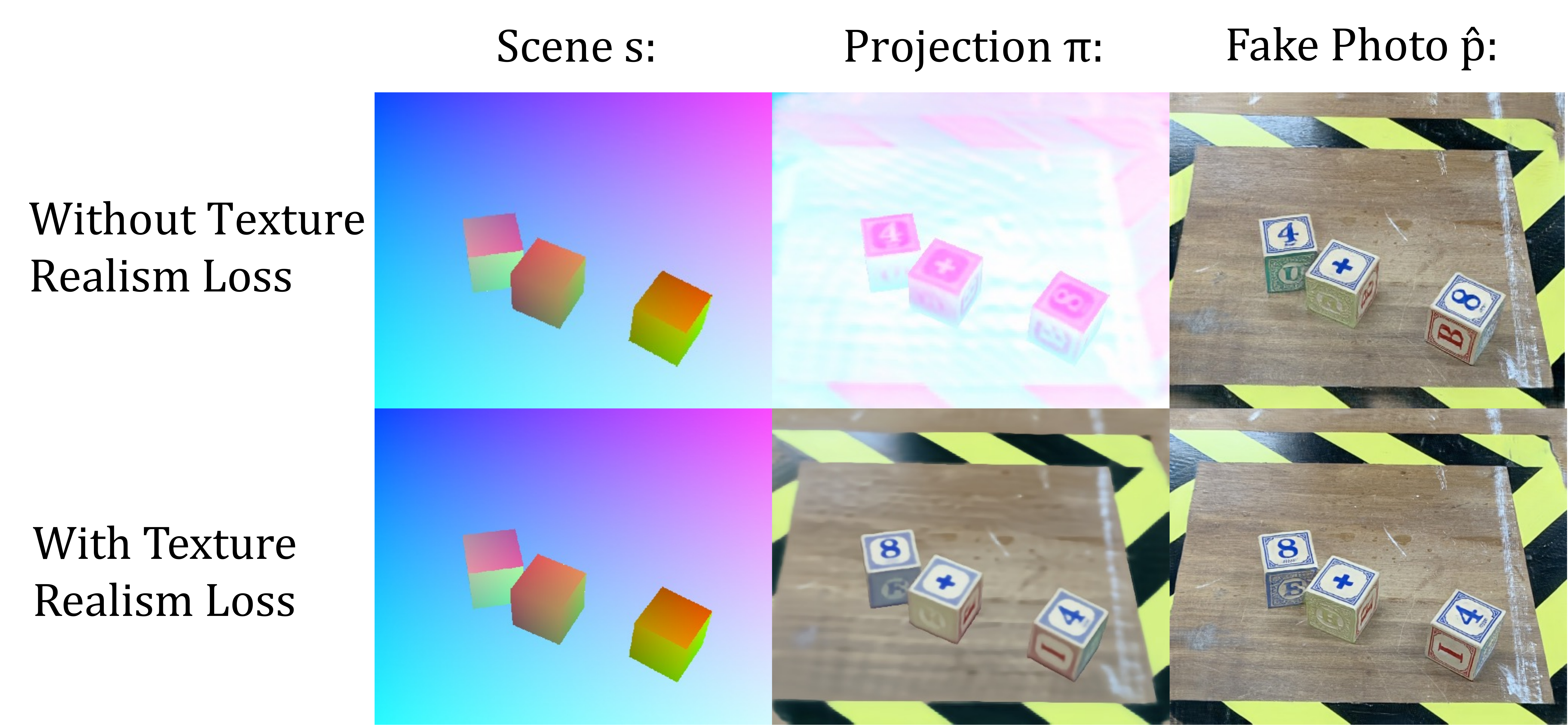}
            \vspace{-5pt}
			\caption{
				The neural neural texture looks more realistic with texture realism loss enabled. Note that the blocks show different letters because of different initializations and the symmetry of a cube - both solutions are valid texture assignments. 
			}
			\label{fig:texture_realism_ablation}
    \end{minipage}
    \vspace{-12pt}
\end{figure}
\vspace{-5pt}
\subsection{Unprojection Consistency Loss}
\vspace{-5pt}
	\label{sec:unprojection_consistency_loss}

		To keep the object surfaces consistent, we impose a pixel-wise ``unprojection consistency loss'' by unprojecting surfaces in fake photos back into the common texture space.
        Given a UVL scene $s$ and its respective fake photo $\ph$, the unprojection $\unprojection$ is obtained from assigning $(r,g,b)$ values at each pixel location $(x, y)$ of $\ph$ to its corresponding $(u, v, l)$ coords given by $s$. 
        For simplicity, we note
        \begin{equation}
            	\unprojection_{[u,v]}^{l}=\expectation\ph_{[x,y]} \quad s.t.~ (u,v,l) = s_{[x,y]},
        \end{equation}
        where the expectation $\mathbb{E}$ means we aggregate multiple $(r,g,b)$ vectors being assigned to the same $(u,v,l)$ coordinate by averaging them.
        In practice, $(u,v,l)$ are real numbers between $[0,1]$, where as each $\unprojection$ have to be rasterized, i.e., represented by $L$ images with a size of $(D\times D\times 3)$, where $L$ is the number of labels and $D\times D$ is the resolution of the unprojection. Therefore, we discretize and scale corresponding $(u,v,l)$ to integers so that each $(r,g,b)$ vector will be assigned to a pixel on $\unprojection$. For the exact implementation, please see our supplementary material.
        
        We obtain $N$ unprojections $\{\unprojection_{[u,v]}^{l,i}\}_{i=1}^{N}$ from a batch of $N$ UVL scenes and fake photos. The unprojection consistency loss is defined as the per pixel-channel standard deviation of $\unprojection$ over the batch
        \begin{equation}
            \ell_{UC} = \frac{1}{L\times D\times D\times 3}\sum_{l,u,v,c}\sigma(\{\unprojection^{l,i}_{[u,v,c]}\}_{i=1}^{N}),
        \end{equation}
        where $\sigma(\cdot)$ stands for the standard deviation function and $c \in \{1\dots 3\}$ is the channel index of $\unprojection$.
		We minimize $\lUC$ to encourage unprojections to be consistent across the batch. Intuitively, if $\lUC$ were $0$, it would mean the object surfaces in translations $\ph$ appear exactly the same in every scene. In addition, we call the mean of unprojections $\mu(\{\unprojection^{l,i}\}_{i=1}^N)$ as ``Recovered Textures'', visualzed in Figure~\ref{fig:first_diagram} and Section~\ref{fig:unprojection_consistency_loss}.

    \vspace{-3pt}
	\subsection{Texture Realism Loss}
	\vspace{-3pt}
	\label{sec:texture_realism_loss}

		To encourage the neural textures to look as realistic as possible, we try to make the projections look like the final output by introducing a ``texture realism'' loss $\lTR$. 
		$\lTR$ is an image similarity loss that makes $\pi$ look like its translation $\hat{p}$.
		\begin{equation}
			\lTR = \similar(\pi,\ph)
		\end{equation}
		$\similar$ is the image similarity from Equation \ref{eq:omega}. Without $\lTR$, $\psi$ can look wildly different each time you train it.
		As seen in the top row of Figure \ref{fig:texture_realism_ablation}, the neural texture looks very unrealistic --- the colors are completely arbitrary.
        By adding texture realism loss $\lTR$, we make the textures $\psi$ more realistic. In practice, this makes TRITON less likely to mismatch the identity of translated objects.

\section{Experiments}\label{sec:results}
In this section, we perform two quantitative experiments: the first measures TRITON's accuracy directly, and the second measures it's sim2real capability in a real-world setting. \emph{Please see the appendix for more interesting experiments!}
\vspace{-3pt}
\subsection{Datasets}
\label{sec:datasetsresults}
\vspace{-3pt}
We constructed two datasets AlphabetCube-3 and RobotPose-xArm to benchmark different image translators in many perspectives.
In our setting, each dataset is composed of two sets of unpaired images: UVL scenes from a simulator and real photographs.
Note that the images in all datasets are unpaired and UVL scenes only rely on rough 3D models of the objects in the scene without need of precisely aligning objects in the simulator to the real world.
We use AlphabetCube-3 for image translation quality evaluation, and RobotPose-xArm for sim2real policy learning evaluation.

\begin{figure}[t]
    \vspace{-20pt}
	\begin{center}
		\includegraphics[width=0.95\textwidth]{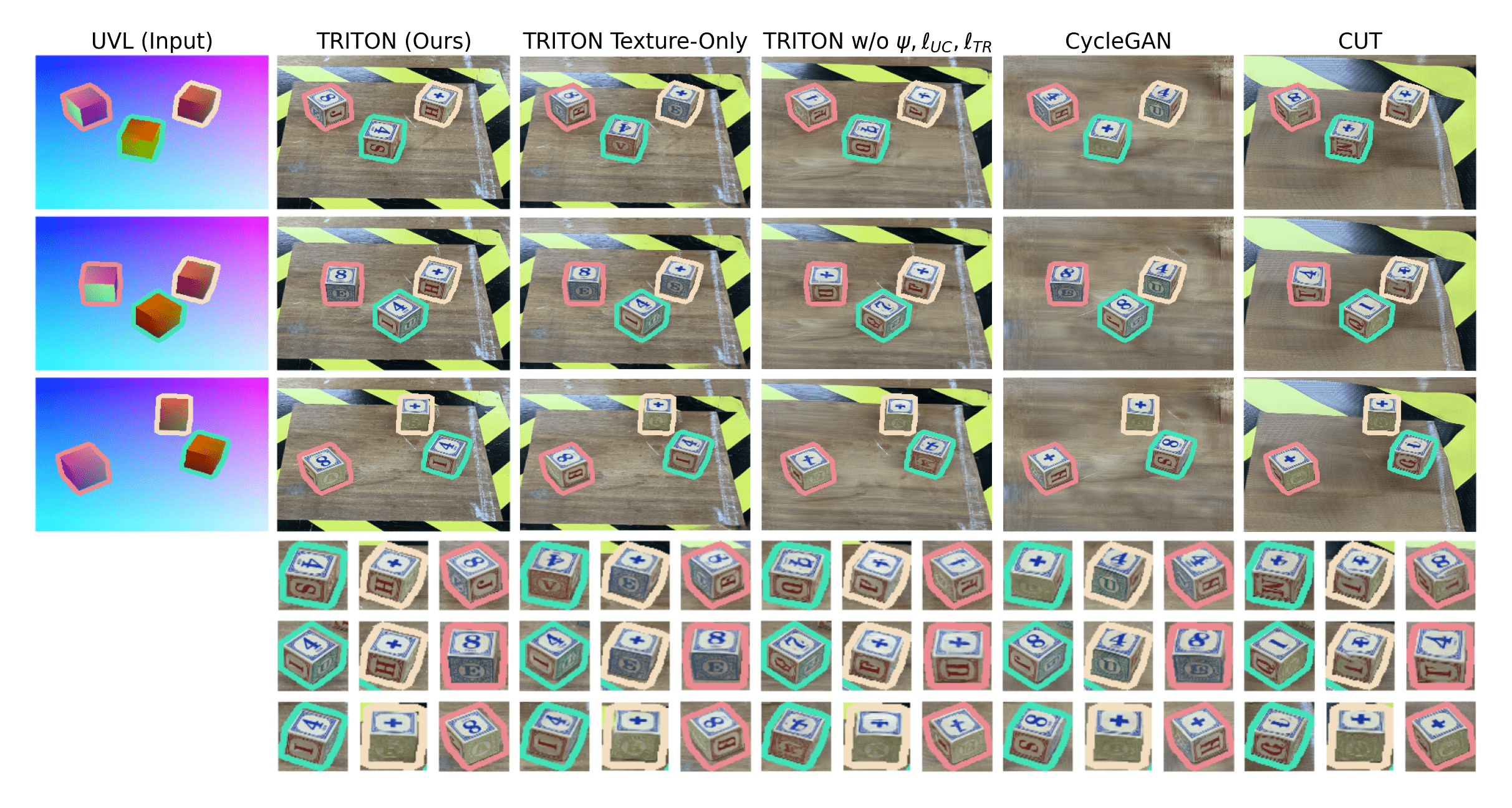}
	\end{center}
	\vspace{-13pt}
	\caption{
	    Image translation comparisons showing how TRITON has better temporal consistency.
	    The first column is the UVL map as the input.
	    The other columns are results using different image translation methods.
		Each row of images shows a different random placement of objects in the scene.
		TRITON consistently outputs high quality results over different object arrangements. 
		``TRITON Texture-Only'' stands for TRITON without two surface consistency losses $\lUC$ and $\lTR$.
		}
		\vspace{-12pt}
		\label{fig:frame_inconsistency_diagram}
	\end{figure}

\subsection{Image Translation Evaluation on AlphabetCube-3}
AlphabetCube-3 dataset features a table with three different alphabet blocks on it.
The dataset contains 300 real photos and 2000 UVL scenes from a simulator.  
Each photo features these three cubes with a random position and rotation.
In this section, we evaluate the image translation accuracy of TRITON and other image translation methods using AlphabetCube-3. In this dataset, $L=4$ because there are four different textures: three textures for the three cubes and one for the table.

Figure~\ref{fig:frame_inconsistency_diagram} shows qualitative results on the dataset.
From Figure~\ref{fig:frame_inconsistency_diagram} we find that TRITON consistently outputs high quality results over different object arrangements.
The textures keep aligned even when the position of blocks change dramatically over multiple scenes (See examples in green rectangles in Figure~\ref{fig:frame_inconsistency_diagram}).
In contrast, though MUNIT~\cite{munit} and CycleGAN~\cite{cyclegan} manage to generate realistic images, the surfaces of the cubes are either consistent over scenes (See examples in the red rectangle) or replicated by mistake (See examples in the orange rectangle).
Also note that the floor background of the outputs from CycleGAN are quite different from the ground truth.
CUT~\cite{cut} fails to generate meaningful images regarding both the foreground blocks and the background.

In quantitative experiments shown in Table~\ref{tab:quantitative}, we manually align the simulator with 14 photos of different real world scenes and generate the UVL maps for translation.
We measured the LPIPS~\cite{lpips} and $l2$-norm between the translated images and the real images using multiple configurations.
`Masked' means we mask out the background (which is the floor in AlphabetCube-3 dataset) and only measure the translation quality w.r.t three foreground blocks.
For the `unmasked' tests, we compare the whole image without any masking instead, and the background contributes much more to the losses due to its larger area.
Similar to the qualitative results, TRITON consistently outperforms other methods under different metrics and configurations.
Further ablations TRITON w/o $\psi, \lUC, \lTR$ and TRITON w/o $\lUC, \lTR$ show the importance of the new losses we introduce.

\begin{table}[thbp]
    \centering
    \setlength{\tabcolsep}{2pt}
    \caption{Quantitative results on AlphabetCube-3. LPIPS~\cite{lpips} and $l2$-norm between the translated images and the real images are reported. We exclude backgrounds in the `Masked' configuration. TRITON consistently outperforms other methods in different metrics and configurations}
    \label{tab:quantitative}
\resizebox{.80\textwidth}{!}{
        \begin{tabular}{lcccc}
        \toprule
        & \multicolumn{2}{c}{Masked} & \multicolumn{2}{c}{Unmasked}  \\
        \cmidrule(lr){2-3} \cmidrule(lr){4-5} & LPIPS ($\times 10^{-1}$) $\downarrow$ & L2 ($\times 10^{-2}$) $\downarrow$  & LPIPS ($\times 10^{-1}$) $\downarrow$  & L2 ($\times 10^{-2}$) $\downarrow$   \\
        \midrule
        CycleGAN & 0.450 &   0.559 &    6.02 &    4.25  \\
        CUT & 0.469 & 0.553  & 5.40  & 6.37 \\
        \midrule
        TRITON w/o $\psi, \lUC, \lTR$ & 0.437   & 0.500  & 4.34  & 4.25  \\
        TRITON w/o $\lUC, \lTR$       & 0.442  & 0.586  & 2.85 & 3.64 \\
        \textbf{TRITON} & \textbf{0.286}  & \textbf{0.479}  & \textbf{1.17}  & \textbf{1.23} \\
        \bottomrule
        \end{tabular}
	}
\end{table}

\vspace{-8pt}
\subsection{Sim2real Transfer for Robot Learning}
\label{sec:robotarm_results}
\begin{figure}[b]
    \centering
    \vspace{-15pt}
        \includegraphics[width=350pt]{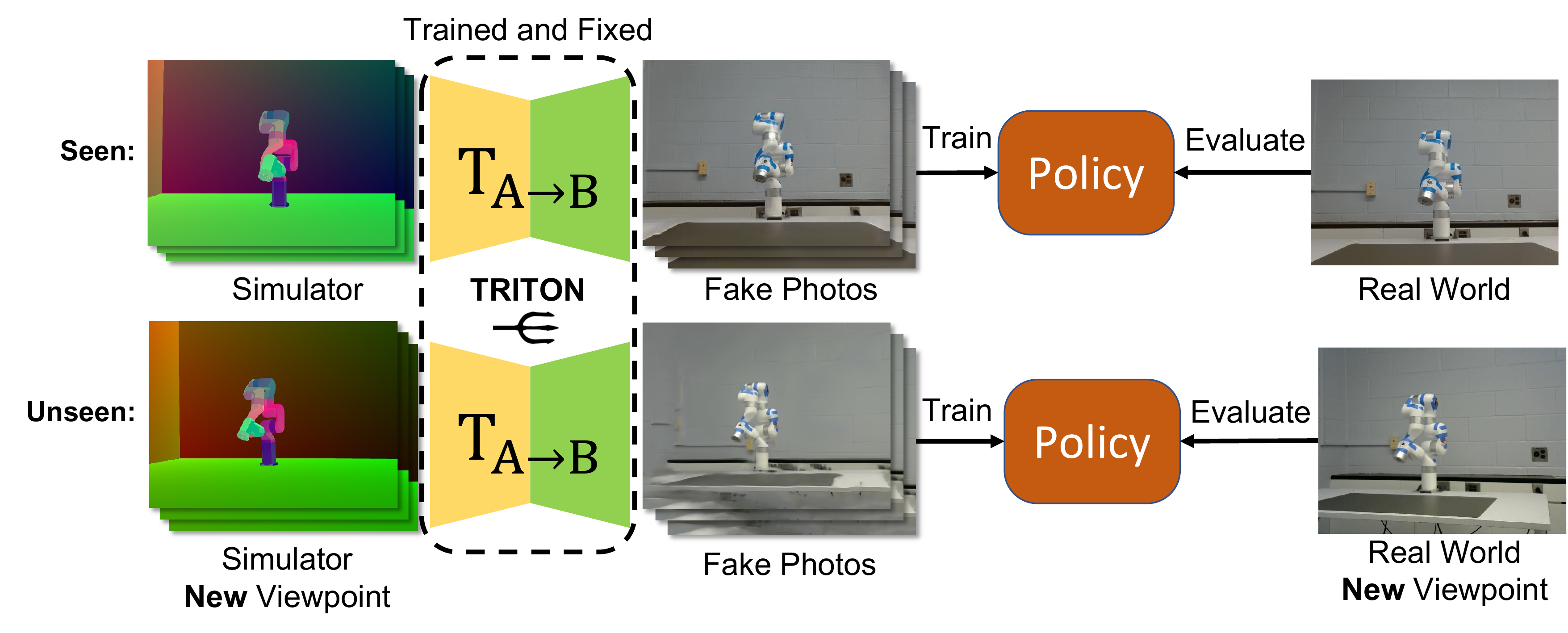}
    \vspace{-10pt}
    \caption{Sim2real framework of using TRITON. We evaluate in two settings: Seen and Unseen. In Unseen, we use a new camera viewpoint to train and evaluate the policy. This new viewpoint is not used for training TRITON. }
    \vspace{-8pt}
    \label{fig:sim2realframework}
\end{figure}
In this section we demonstrate how TRITON improves sim2real transfer for robot policy learning. To this end, we first train TRITON on a dataset consist of unpaired UVL scenes from a robot simulator (Gazebo) and photos from the real robot. Once TRITON is trained, we learn the robot policy while only utilizing the simulator. The input of the policy is the fake photo $\ph$ translated from the simulated UVL scene using TRITON's translator $T_{A\to B}$. Finally, the trained policy is directly evaluated on the real robot, taking real photos $p$ as input.

Ideally, a good image translation model makes $\ph$ similar to $p$, so that the policy trained on synthesized photos will transfer to real domain seamlessly and will have better performance. All our policies are learned with zero real-world robot interaction with the environment.

\textbf{Task} We use a robot pose imitation task for our experiments. Given the input of a photo of real robot pose, the policy outputs robot controls to replicate that target pose. We measure the angular error between the replicated and target poses as our evaluation metric:$\sqrt{\sum_{j} (\hat{a}_j - a_j)^2}$,
where $\hat{a}_j$ and $a_j$ are replicated and target joint angles respectively, and $j$ is the joint index. We use an xArm robot which has seven joints. We also attached patterns and tapes on the robot to make its texture more challenging to model.
\textbf{Robot policy and Baselines}
Since the sim2real formulation allows benefiting from abundant training data using the simulator, we use behavioral cloning to train a good robot policy.
The policy network is a CNN similar to~\cite{dqn}.
We compare TRITON against two baseline image translation methods: CycleGAN~\cite{cyclegan} and CUT~\cite{cut}, by replacing TRITON with each baseline method respectively in the above pipeline. The robot policy learning method is same for all the methods.

\textbf{Evaluation Settings} We introduce two main evaluation settings, \textbf{Seen} and \textbf{Unseen}. In the \textbf{Seen} setting, the policy is trained and tested using the same camera viewpoint that is used during TRITON (or baseline) training. Whereas in \textbf{Unseen} setting, we introduce a camera at a new viewpoint for training and testing the policy. That is, the image translator has to generate the fake photos out of its training distribution (seen viewpoints), which becomes a challenging task for the translator. We also vary input image resolutions to show the power of photo-realistic images in higher resolutions.

\begin{figure}[tb]
    \centering
    \vspace{-5pt}
    \includegraphics[width=0.95\textwidth]{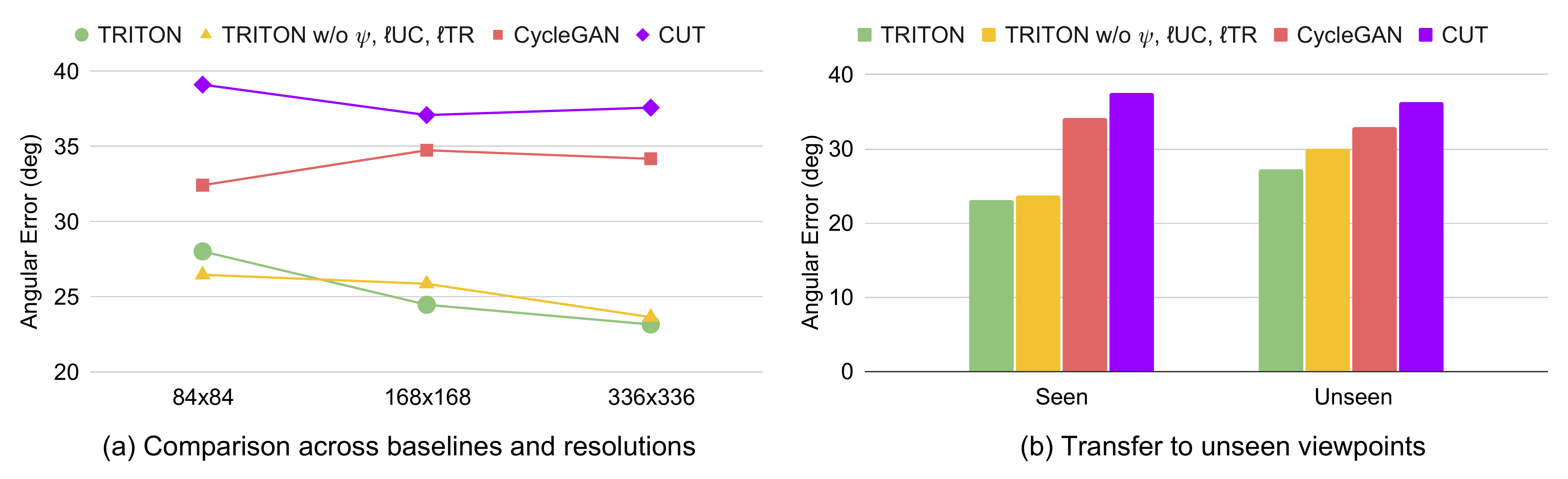}
    \vspace{-10pt}
    \caption{Evaluation results of sim2real robot learning. We compare TRITON against two baselines and one TRITON variant w/o $\psi,\ell_{UC},\lTR$ across 3 different input image resolutions. TRITON outperforms CycleGAN~\cite{cyclegan} and CUT~\cite{cut} consistently across (a) all resolutions and (b) unseen viewpoints. Results in (a) are from Seen setting only. Results in (b) are from 336x336 resolution. }
    \label{fig:sim2real}
    \vspace{-5pt}
\end{figure}

\textbf{Results} Figure~\ref{fig:sim2real} shows the evaluation results of the sim2real transfer. Both Seen and Unseen evaluations show that TRITON outperforms baselines consistently. In Figure~\ref{fig:sim2real}(a), we find that TRITON continuously improves from low to high resolutions due to its ability to generate more photo-realistic images compared to the baselines.
From Figure~\ref{fig:sim2real}(b), we confirm that TRITON outperforms the others also in the Unseen setting, showing that TRITON learns image translations that better generalize to new viewpoints.
TRITON outperforms TRITON w/o $\psi,\ell_{UC},\lTR$ consistently except at the lowest resolution. This again shows the effectiveness of $\psi$, $\ell_{UC}$ and $\lTR$ to generate photo-realistic images, and such high quality images are important in higher resolutions for sim2real transfer.

In Appendix in the supplementary material, we provide more experimental results.

\vspace{-9pt}
\section{Limitations}
\label{sec:Limitations}
\vspace{-3pt}
TRITON relies on 3D models of objects in order to generate photo-realistic images, making it less applicable when the geometry of an environment is unknown. Acquiring such 3D models could be challenging in the wild, especially in robots allowed to roam the world. TRITON is not able to handle transparent objects, because the UVL format is opaque. Generated shadows, though visually realistic, are not physically accurate. Moreover, the probability of incorrectly matching textures to objects increases as the number of object classes increases.

\acknowledgments{We thank Srijan Das and Kanchana Ranasinghe for valuable discussions. This work is supported by Institute of Information \& communications Technology Planning \& Evaluation (IITP) grant funded by the Ministry of Science and ICT (No.2018-0-00205, Development of Core Technology of Robot Task-Intelligence for Improvement of Labor Condition. This work is also supported by the National Science Foundation (IIS-2104404 and CNS-2104416).}

\clearpage

\bibliography{example}  %

\newpage

\input{appendix.tex}

\input{rebuttal.tex}

\end{document}

%% file: appendix.tex
\appendix

\section*{Appendix} 

\section{Algorithmic Details}
\subsection{Neural Neural Textures: Why use them? An Ablation}
    \label{sec:appendix_neural_tex}
	In Section \ref{sec:neural_tex}, we mentioned that neural neural textures learn better than raster neural textures. 
	TRITON can be ablated to use raster textures. Let's call this version ``Raster TRITON''. 
	Raster TRITON is extremely sensitive to the resolution of its neural texture, and we found that it can be numerically unstable if that resolution is too high.
	In comparison, regular TRITON's neural neural textures do not have a specific resolution: they are continuously defined over the UV domain using Fourier feature networks.

	\begin{figure}[H]
		\begin{center}
			\includegraphics[width=400pt]{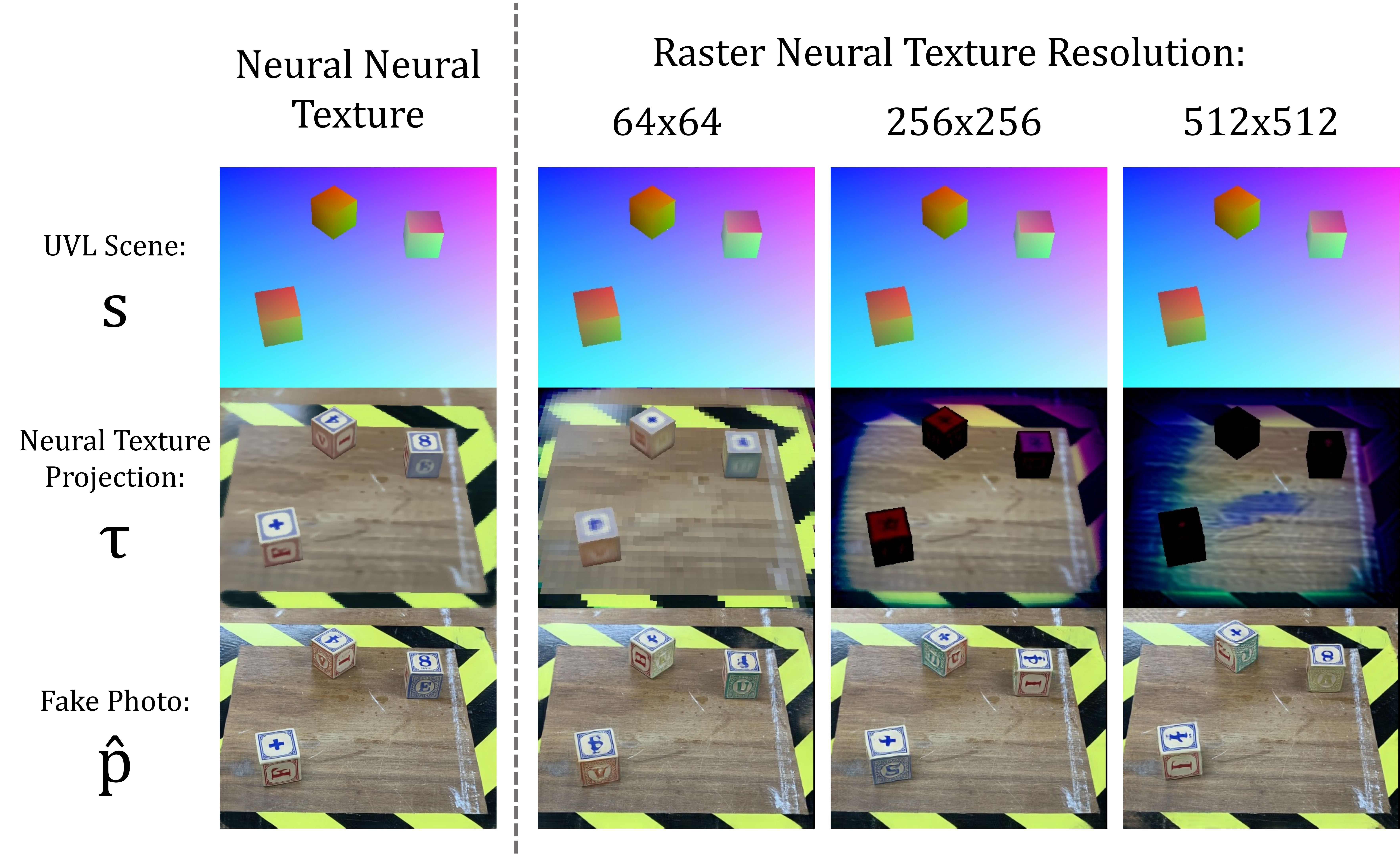}
		\end{center}
		\caption{
			The resolution of the raster neural texture seriously impacts performance in ``Raster TRITON''. Each column has a different neural texture resolution.
			The neural texture only looks realistic when the resolution is low.
		}
		\label{fig:raster_texture_resolution_comparisons}
	\end{figure}

    In Figure \ref{fig:raster_texture_resolution_comparisons}, we observe that the output quality of Raster Triton degrades as the resolution of the neural texture increases, requiring a very low resolution to give a proper output. However, a low resolution neural texture isn't capable of capturing much detail. By comparison, the neural neural texture is able to both capture a high level of detail and learn a realistic texture.
    
	Here's our explanation for this phenomenon.
	With raster-based textures, only the side of an object that currently is seen in a view is allowed to change during each update, because the gradient can not be propagated into pixels which are not seen.
	If this means changing overall brightness of an object for example, the brightness of that object can not be changed over the entire object until every side has been seen - which will not happen in a single iteration, since the batch size is limited.

	In addition to not propagating the gradient to unseen sides of objects, it also cannot propagate to texels in-between the UVL values seen in a scene image.
	When the raster neural texture is too large, aliasing effects occur: if you have a raster texture with very high resolution, the loss gradient is less likely to be passed to a given texel because the chance that a given UV value in a scene will be rounded to that pixel's coordinates is very small.
	In practice, this makes large raster neural textures unstable and limits us to using small amounts of detail.
	With neural neural textures, this aliasing problem is mostly mitigated, because when a certain texture receives a gradient at particular UV coordinates, the areas of the texture between those coordinates are also changed.

\subsection{Unprojection Consistency Loss: More Details}

	Here, we give more details about unprojection consistency loss $\lUC$.

	The exact equation for $\lUC$ is as follows:
	to formally define $\lUC$ we define unprojections $\unprojection^{1...N,1...L}$ and mean unprojections (aka recovered textures) $\bar{\unprojection}^{1...L}$ 
	with each $\unprojection^{n,i}\in\real^{3\times128\times128}$:
	\begin{multline}
		\unprojection_{c,U,V}^{n,i}=\expectation\ph_{c,x,y}^n 
		\eqand \bar{\unprojection}^i_{c,u,v}   =   
			\frac{1}{N} \lsum_{n=1}^N\unprojection^{n,i}_{c,u,v} 
		\eqand \\
		\lUC=\frac{\lsum_{i=1}^L\lsum_{c=1}^3\lsum_{U=1}^{128}\lsum_{V=1}^{128}
		\sqrt[]{\frac{1}{N}\lsum_{n=1}^N\left(\bar{\unprojection}^i_{c,u,v}-\unprojection^{i,n}_{c,U,V}\right)^2}}
		{3\cdot128\cdot128\cdot L}
	\end{multline}
	where $U=\lfloor128u\rfloor$ and $V=\lfloor128v\rfloor$ and $i=I(l)$ with
	$u=s^n_{1,x,y}$, $v=s^n_{2,x,y}$, $l=s^n_{3,x,y}$
	for all 
	$n \in \{1...N\}$ , 
	$i \in \{1...L\}$ , 
	$x \in \{1...W\}$ , 
	$y \in \{1...H\}$ , 
	$U \in \{1...128\}$ , 
	$V \in \{1...128\}$.

	\bigskip
	
\begin{figure}[H]
		\begin{center}
			\includegraphics[width=400pt]{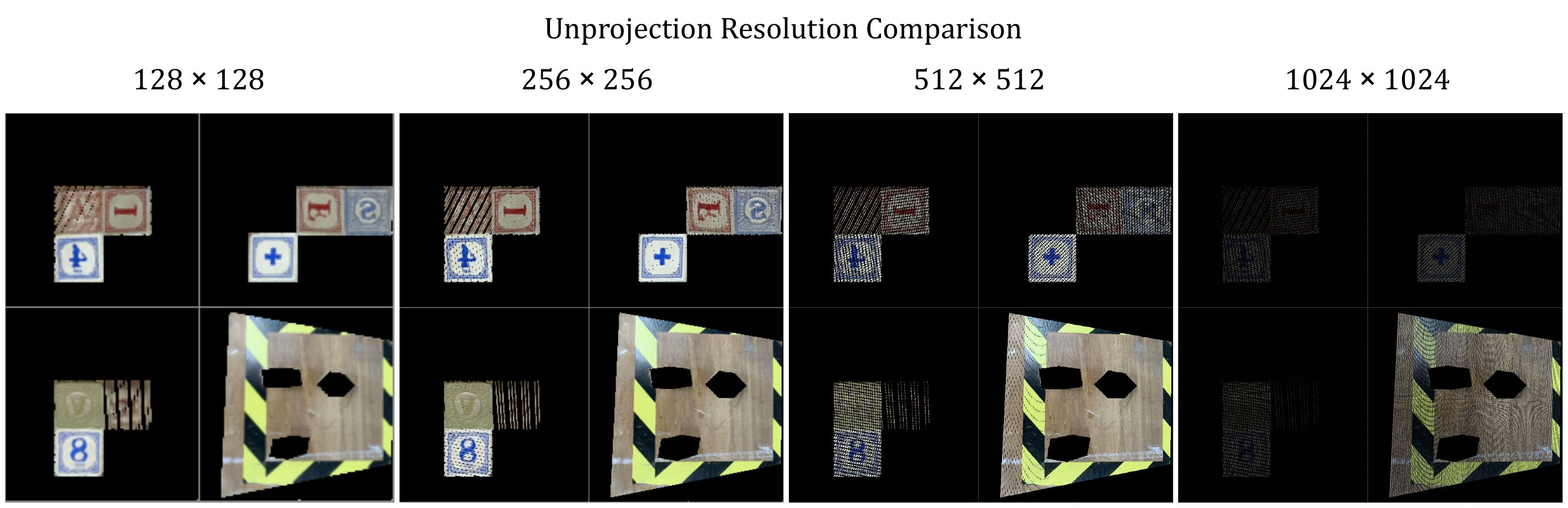}
		\end{center}
		\caption{
			Here, we unproject a single fake photo $\ph^i$. The resolution of the unprojection matters for unprojection consistency loss. The larger it is, the more precise the alignment will be but the less likely a given UV value is to be assigned a loss greater than 0, creating numerical instability and slowing down learning. 
			When the unprojection resolution is too high, only a few areas of the neural texture will receive a gradient during backpropogation. In practice we use the resolution $128\times128$.
		}
		\label{fig:unprojection_resolution_comparison_weetibeeto}
	\end{figure}

	The hyperparameter 128 we described in Section \ref{sec:unprojection_consistency_loss} refers to the resolution of the unprojections used to calculate $\lUC$.
	In Figure \ref{fig:unprojection_resolution_comparison_weetibeeto} we see that if we were to set it higher, the gradient wouldn't affect the neural texture as densely.

\subsection{Training Procedures}
\label{ref:training_procedures}
	In this section we detail the training procedures used to create results in Section \ref{sec:results} from Triton, CycleGAN, CUT,
		TRITON w/o $\lUC,\lTR$ (aka TRITON with neither unprojection consistency loss nor texture reality loss), and
		TRITON w/o $\psi, \lUC,\lTR$ (aka TRITON without textures, and therefore also without unprojection consistency loss or texture reality loss).

	\subsubsection{TRITON}
	\label{par:triton}
	We train TRITON for 200,000 iterations on all datasets. Each of these datasets is comprised of two sets of \emph{unpaired} RGB images: one containing simulated UVL images, and the other containing real photographs. This takes about 36-48 hours on an NVIDIA RTX A5000. 
	The dimension of each input scene is defined by hyperparameter height $H$; the width is scaled to match the aspect ratio of the original input.
	To avoid running out of video memory, we randomly crop each input to a $256\times256$ subset and run TRITON on that cropped input.
	We use batch size 5 during training.
	
	We found that the best way to train TRITON is to start with smaller $H$ values and progressively increase that value during training.
	For the first 50,000 iterations we use $H=320$, then for the next 50,000 iterations $H=420$, then for the last 100,000 iterations $H=\text{(the height of the original input scene)}$.

	Like in \cite{surgical_video_translation}, we use three Adam optimizers: two for the translation module's encoders and decoders, and another for the neural texture.
	For the translation module's optimizers, we use learning rate $\num{1e-4}$, and for the neural neural texture we use learning rate $\num{1e-3}$. 
	Every 100,000 iterations all learning rates are halved.

	During evaluation, we render the neural neural texture onto a raster grid of pixels. Because the neural neural texture has a 2d manifold, it can be closely approximated by an RGB image. Using this method lets us avoid evlauating $\psi$'s fourier feature network on every frame by using a raster image as a lookup table. This decreases video memory usage and speeds up calulations during inference.

	\subsubsection{TRITON w/o $\lUC,\lTR$}
	\label{par:triton_without_luc}
	This ablation is also known as ``Texture-Only'' TRITON, because it has a learnable texture $\psi$ but no consistency losses.
	Its trained the same way that we train TRITON normally as discussed above in \ref{par:triton}, except the surface consistency losses $\lUC$ and $\lTR$ are omitted.
	Effectively, it performs better than MUNIT-like TRITON (aka TRITON w/o $\psi, \lUC,\lTR$, discussed in \ref{par:munit_triton})

	\subsubsection{TRITON w/o $\psi, \lUC,\lTR$}
	\label{par:munit_triton}
	This ablation is very similar to MUNIT, as TRITON is based on MUNIT and this ablation has neither learnable texture nor consistency losses.
	Like in subsection \ref{par:triton_without_luc}, this ablation shares the same training procedures as TRITON.
	
	\subsubsection{Raster TRITON}
	\label{par:raster_triton_baseline}
	``Raster TRITON'' refers to a variant of TRITON that uses a discrete grid of pixels for its neural texture instead of a neural neural texture $\psi$. More details about this type of ablation are discussed in section \ref{sec:appendix_neural_tex}, and depicted in Figure \ref{fig:raster_texture_resolution_comparisons}.
    In all tests involving ``Raster TRITON'', we use a neural texture with a resolution of $128\times128$ texels.

	\subsubsection{CycleGAN}
	\label{par:cyclegan}
	We use the original implementation of CycleGAN, and stick to the reccomended 200 epochs, as it tends to overfit if you go further. 
	For our tests, we run CycleGAN on multiple different scales of the input scenes, with input sizes $286\times286$ (CycleGAN's default), $320\times320$, $420\times420$ and $512\times512$. 
	After translation, we stretch the image into the input's aspect ratio.
	We also set $\lambda_{identity}=0$, meaning we disable the identity loss.
	The reported results are the calculated using the best results among these resolutions.
	This dramatically improves CycleGAN's performance when translating UVL scenes to photographs, as this loss was built with the assumption that some parts of the input should not be changed during translation (which is not true in our case).
	
	\subsubsection{CUT}
	\label{par:cut}
	We use the original implementation of CUT, and stick to the recommended 400 epochs, as it tends to overfit if you go further. 
	Like with CycleGAN in \ref{par:cyclegan}, we run CUT on multiple different resolutions of the input scenes, with input sizes $286\times286$ (CUT's default), $320\times320$, $420\times420$ and $512\times512$. 
	After translation, we stretch the image into the input's aspect ratio.
	The reported results are the calculated using the best results among these resolutions.
	
	\subsubsection{MUNIT}
	\label{par:munit_baseline}
	We mainly follow the official implementation of MUNIT~\cite{munit}, except that we use a fixed style code suggested by \cite{surgical_video_translation}, because they demonstrated doing so yielded better temporal coherence.

\section{More Dataset Details}

Because of the unique task of our network, we have created our own datasets. Here are some more details about various datasets used in the paper. 

\subsection{AlphabetCube-3}
AlphabetCube-3 dataset features a table with three different alphabet blocks on it.
The dataset contains 300 real photos and 2000 UVL scenes from a simulator. All photos are taken from an iPhone XS Max. 
Each photo features these three cubes with a random position and rotation where only the z-axis is rotated. The positions are bounded within the striped tape shown on the table. In this dataset, $L=4$ because there are four different textures: three textures for the three cubes and one for the table.

\subsection{RobotPose-xArm}
We collected 891 different robot arm poses in the RobotPose-xArm dataset. Each pose was collected from 3 different camera views. The dataset was collected in pairs of simulated and real images -- for evaluation purposes -- resulting in 891*3 simulated images and 891*3 real images in total. In this dataset, L=2, which are the textures of the background and the robot arm.

\section{More Results}
	In this section, we highlight some more interesting results and observations from the experiments conducted in the main paper. 

	\subsection{Unprojection Comparisons}

		In this section we expand on the results in Subsection \ref{sec:datasetsresults} by showing the mean unprojection (aka recovered texture) for different image translation algorithms. With the 14 UVL images and corresponding fake photos (or photos) as calculated in \ref{sec:datasetsresults}, we average their unprojections.
		If a translation algorithm is consistent, these unprojections should align well and the average $\bar{\omega}$ emphould not be blurry.

		\begin{figure}[H]
			\begin{center}
				\includegraphics[width=400pt]{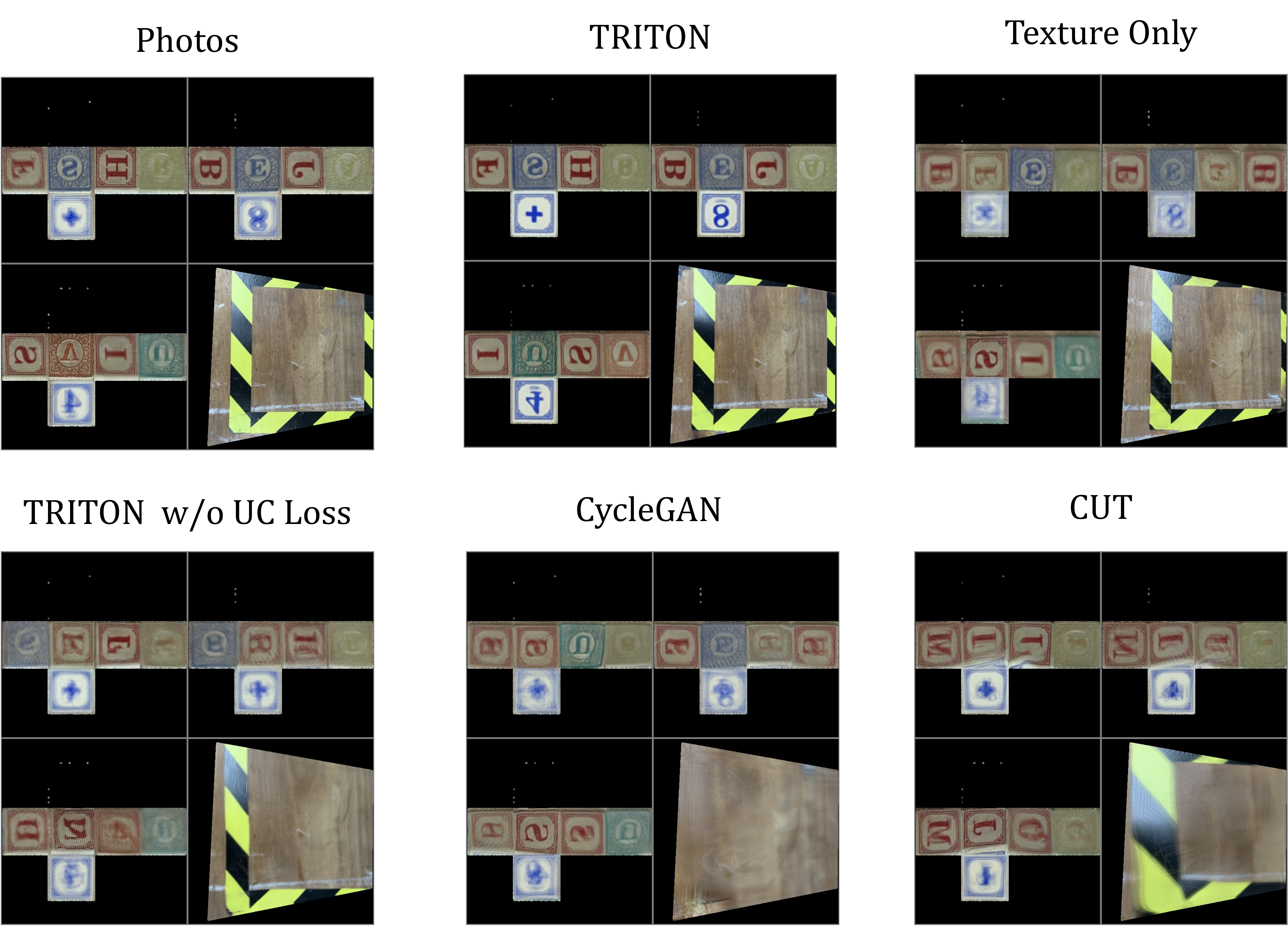}
			\end{center}
			\caption{
				In this figure, we compare the recovered textures $\bar{\omega}$ (aka mean unprojections) of the 14 labeled images between different algorithms, and compare it to a ground truth. In this diagram, we align all the unprojections to make them easier to compare. Note how TRITON is more crisp and matches the unprojected photographs better than the other algorithms, which are blurrier and have the wrong colors and letters on each side of the alphabet blocks. CycleGAN and CUT for example incorrectly duplicate letters between alphabet blocks.
			}
			\label{fig:unprojection_resolution_comparison}
		\end{figure}

\pagebreak
	\subsection{Videos}

		In this section we have links to animations that help demonstrate various aspects of TRITON.
		If you are viewing this document on a computer, click a thumbnail to view the video.

		\begin{figure}[H]
			\begin{center}
			\href{https://youtu.be/kecK_cJgLT8}{
					\includegraphics[width=300pt]{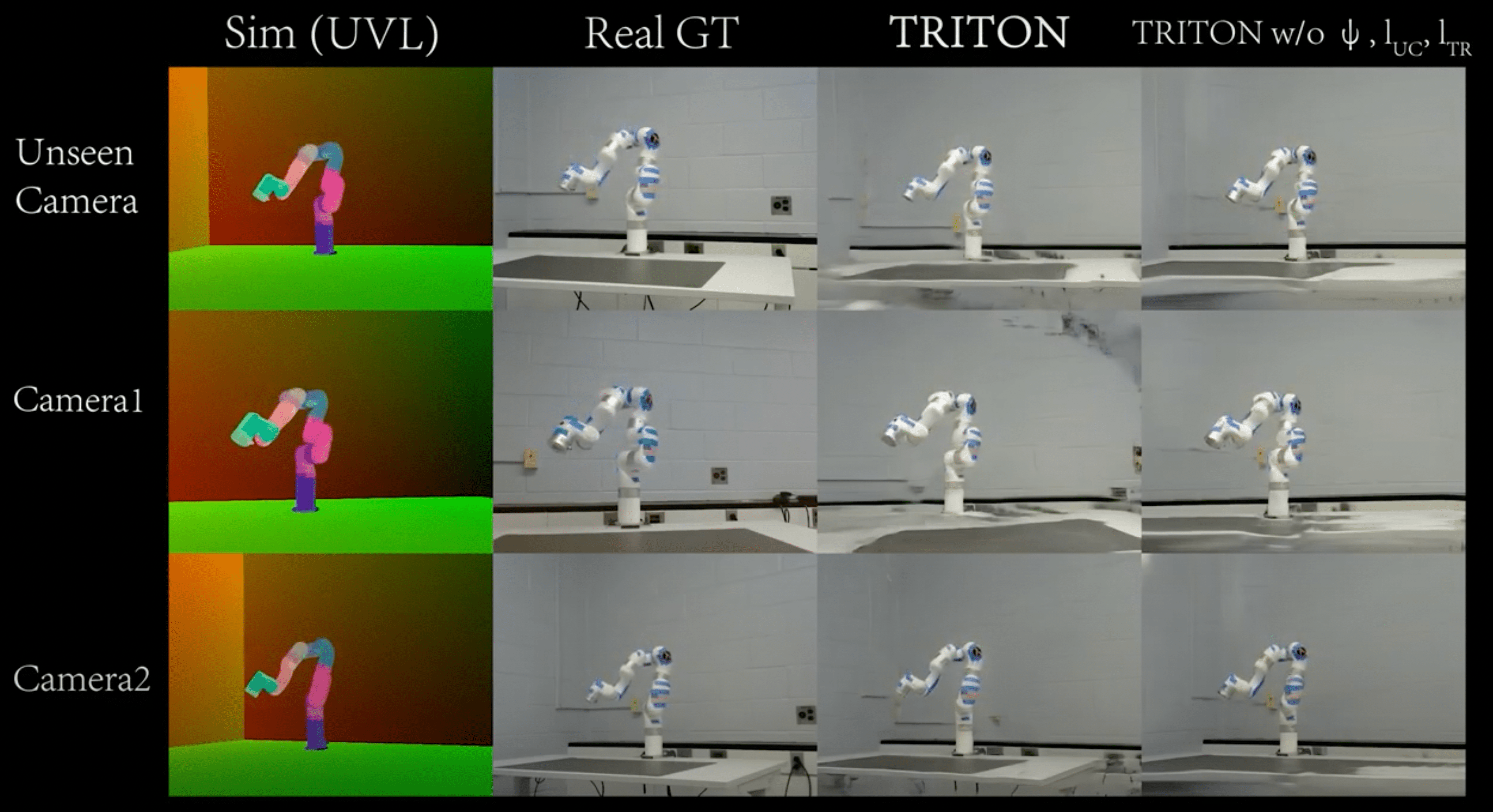}
					}
				\end{center}
			\caption{
				This animation shows a side-by-side comparison of different camera viewpoints of the results discussed in Section \ref{sec:robotarm_results}, as well as the inputs, ground truth video, TRITON and ablation videos.
				Url: \href{https://youtu.be/kecK_cJgLT8}{youtu.be/kecK\_cJgLT8}
			}
			\label{fig:robotarm_anim}
		\end{figure}

		\begin{figure}[H]
			\begin{center}
			\href{https://youtu.be/-GSixT4shxY}{
					\includegraphics[width=300pt]{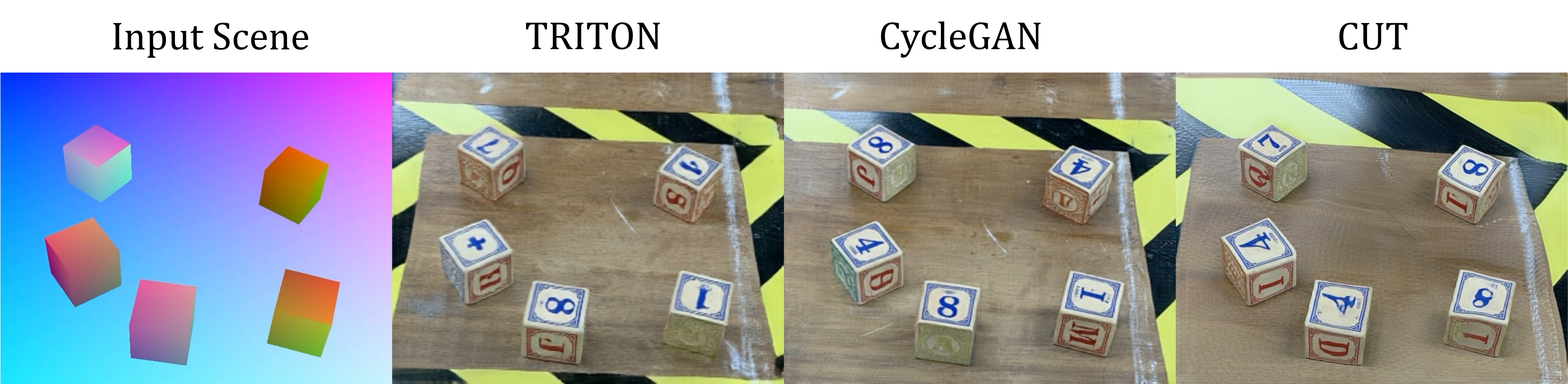}
					}
				\end{center}
			\caption{
				This animation shows an animation of various algorithms on a dataset with 5 alphabet blocks moving around. 
				When watching, note how the numbers on the top of the cubes in both CycleGAN and CUT change over time, while with TRITON they stay the same. If you enjoy this animation, please also see Figure \ref{fig:robot_five_animation}.
				Url: \href{https://youtu.be/-GSixT4shxY}{youtu.be/-GSixT4shxY}
			}
			\label{fig:algocomparisonvideo}
		\end{figure}

		\begin{figure}[H]
			\begin{center}
			\href{https://youtu.be/0t0xiVS_8D0}{
					\includegraphics[width=300pt]{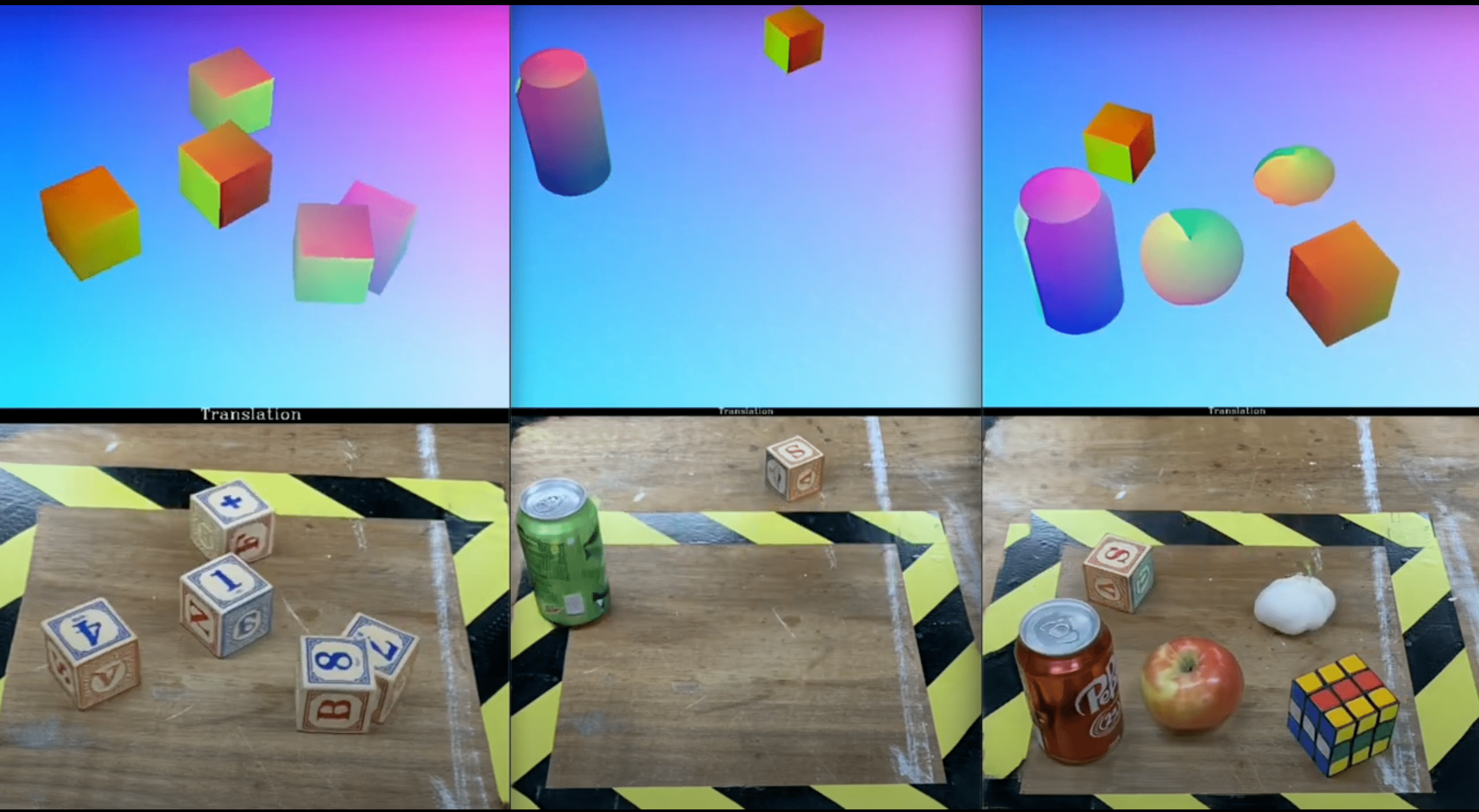}
					}
				\end{center}
			\caption{
				This video shows the results of TRITON applied on three of the datasets shown in Figure \ref{fig:first_diagram}.
				The top row shows the input scenes $s$, and the bottom row shows the fake photographs $\ph$.
				Url: \href{https://youtu.be/0t0xiVS_8D0}{youtu.be/0t0xiVS\_8D0}
			}
			\label{fig:algocomparisonvideo}
		\end{figure}

%% file: rebuttal.tex
\pagebreak
\section{Additional Experiments} 
In this section of the appendix, we conduct additional experiments beyond that of the main paper. In particular, we train a robot to locate and move to objects using the RobotFive dataset, shown in section \ref{sec:Reacher}. We also use synthetic dataets to measure TRITON's performance on deformable objects in section \ref{sec:american_flag} and on rigid objects in \ref{sec:threesynth}.
\label{sec:RebuttalExperiments}
\subsection{Robot Manipulator Experiment - Reacher} 
\label{sec:Reacher}
\begin{figure}[H]
		\begin{center}
			\includegraphics[width=400pt]{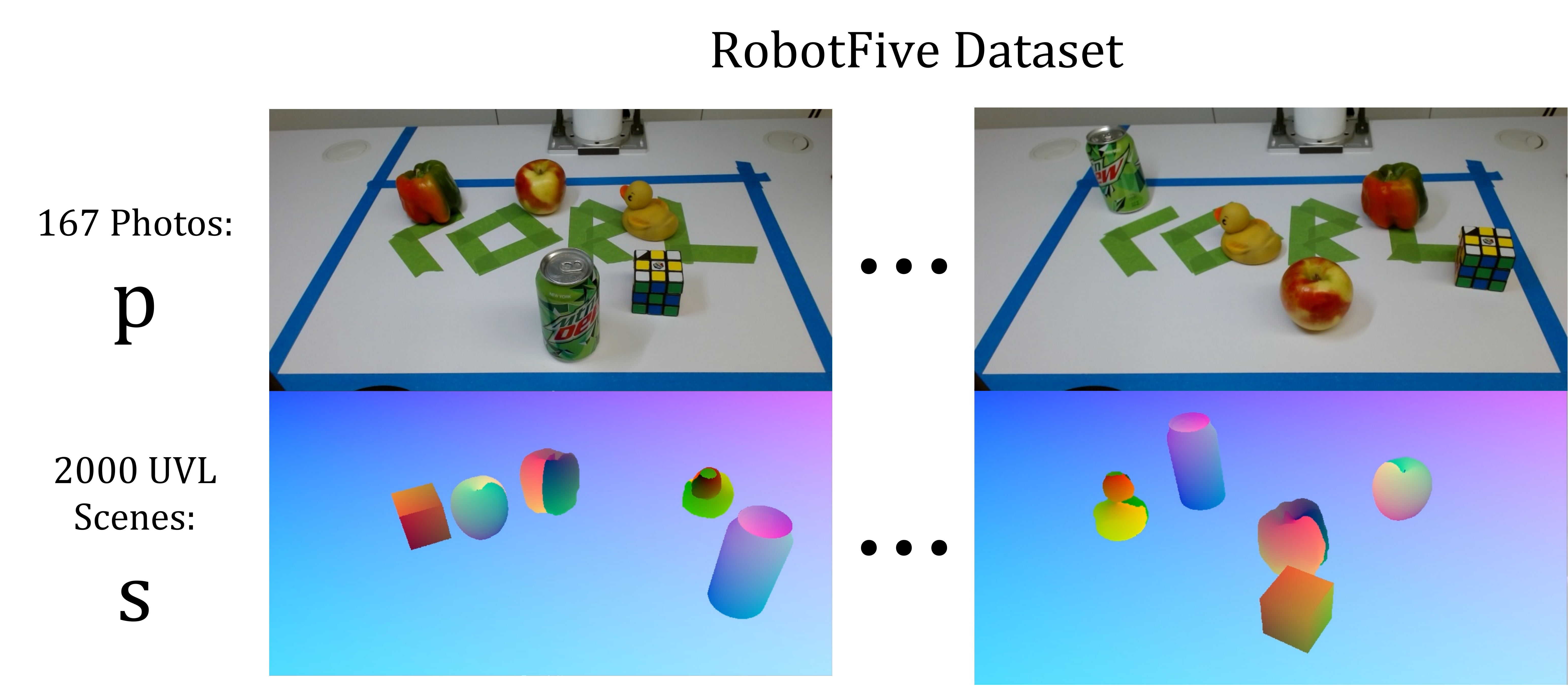}
		\end{center}
		\caption{
			This figure shows some random training samples in the RobotFive dataset.
		}
		\label{fig:robot_five_dataset}
	\end{figure}

    \paragraph{Summary} In this experiment, we conduct real-world robot experiments with behavioural cloning in a sim2real setting.
    Inspired by \cite{li2022does}, the robot takes images from a fixed RGB camera as inputs and the goal is to localize and reach five different objects on a table.
    We learn the policy in a simulation using the image frames translated with TRITON (and other baselines). Then we evaluate them in the real world using actual photographs as input (See Figure~\ref{fig:robot_five_dataset}). 
    By evaluating the learned policies, we can compare the usefulness of TRITON with its various baselines. 
    We find that TRITON produced the most generalizable results, and improved the accuracy of our reacher task (Table~\ref{tab:robotreachexperiments}). %

    \paragraph{Dataset}
    The RobotFive dataset consists of 167 photographs and 2000 UVL scenes, all of which are unpaired. It includes 5 objects that are moved around the table: a pepper, a Rubik's cube, an apple, a rubber duck and a can of soda. Since there are five objects and one table, there are a total of 6 different labels: $L=6$. The height of the images are all 512 pixels. 
    
	\paragraph{Behavioural Cloning}
	Similar to \cite{tossingbot}, we maintain a policy network that takes one single RGB image as input and outputs a heatmap for each object simultaneously, where the values in the heatmap reflect the confidence of an object's existence.
	Inverse kinematics is used to drive the robot to the location with the highest value in the heat map.
	Our network architecture is shown in Table~\ref{tab:robotbcarch} and all the activation functions are ReLU.
	\begin{table}[h]
       \centering
       \caption{Architecture of behavioural cloning policy.}
       \label{tab:robotbcarch}
        \begin{tabular}{c|ccccc}
        \toprule
            Layer number & Type & Kernel size & Output channel & Stride & Padding \\
            \midrule    
            0 & Conv & 3 & 32 & 1 & 1\\
            1 & Maxpooling & 3 & 32 & 2 & 1\\
            2 & Residual block & 3 & 64 & 1 & 1\\
            3 & Maxpooling & 3 & 64 & 2 & 1\\
            4 & Residual block & 3 & 64 & 1 & 1\\
            5 & Upsampling(2x) & - & - & - & -\\
            6 & Residual block & 3 & 32 & 1 & 1\\
            7 & Upsampling(2x) & - & - & - & -\\
            8 & Conv & 1 & 5 & 1 & 0\\
        \bottomrule
        \end{tabular}
   \end{table}
   
	For each channel of the output, we select the pixel where the target object is located as positive and randomly select other $20$ pixels from the whole channel as negative samples. 
	A cross entropy loss is applied to $1 + 20$ pixels.
	
	The network is trained by an Adam optimizer with a learning rate of $10^{-5}$ and a batch size of $8$ for $10$ epochs and the input image is resized to $128 \times 64$ to save computation. 

    \paragraph{Quantitative Results}
    To quantitatively evaluate the results, we manually labeled the positions of all five objects in ten real images in the world coordinates.
    Then we measure the distance between the outputs of the network and the ground-truth.
    Each translation method is trained with multiple random seeds and the mean and standard deviation of error distance are reported in Table~\ref{tab:robotreachexperiments}. TRITON achieves best performance with a significantly lower mean error distance. 
    \begin{table}[h]
       \centering
       \caption{Behavioural cloning results, TRITON achieves best performance with a significant lower error distance.}
       \label{tab:robotreachexperiments}
        \begin{tabular}{lccc}
        \toprule
            Method & Mean distance $\downarrow$ & Standard deviation & Number of runs \\
        \midrule
            {CUT}  & 17.97 & 3.24 & 3\\
            {CycleGAN} & 19.26 & 4.68 & 4\\
            {MUNIT}  & 20.77 & 7.18 & 3\\
            {Raster TRITON} & 16.51 & 1.73 & 4\\
            {TRITON w/o $\psi, \lUC, \lTR$} & 18.65 & 2.78 & 3\\
            \midrule
            {\textbf{TRITON}} & \textbf{11.59} & 6.51 & 4\\
        \bottomrule
        \end{tabular}
   \end{table}

    \paragraph{Qualitative Results}
    
\begin{figure}[H]
    	\begin{center}
    			\href{https://youtu.be/JXNLwWoAlgQ}{
    		\includegraphics[width=400pt]{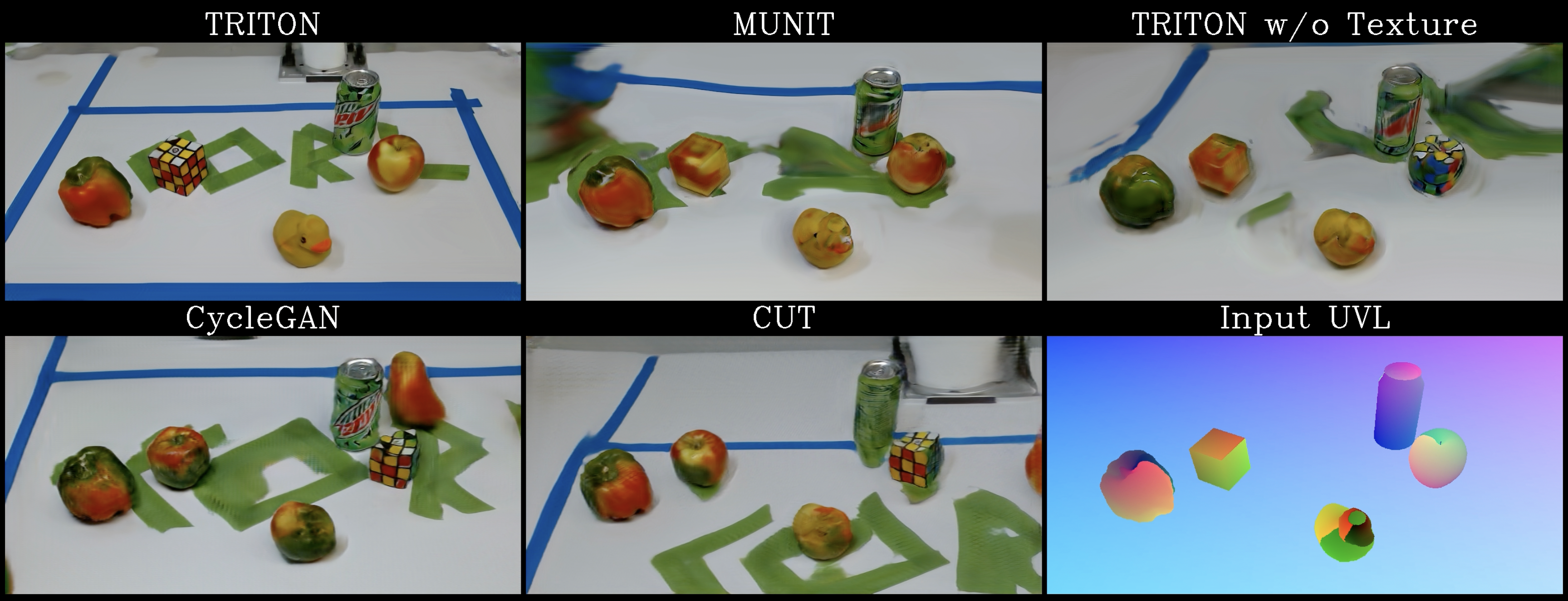}
    		}
    	\end{center}
    	\caption{
    		This animation showcases the results of TRITON and various baselines on the RobotFive dataset. To watch, click the image or visit  \url{https://youtu.be/JXNLwWoAlgQ}.
    	}
    	\label{fig:robot_five_animation}
    \end{figure}

\begin{figure}[H]
    	\begin{center}
    			\href{https://youtu.be/IwAs420hY6A}{
    		\includegraphics[width=400pt]{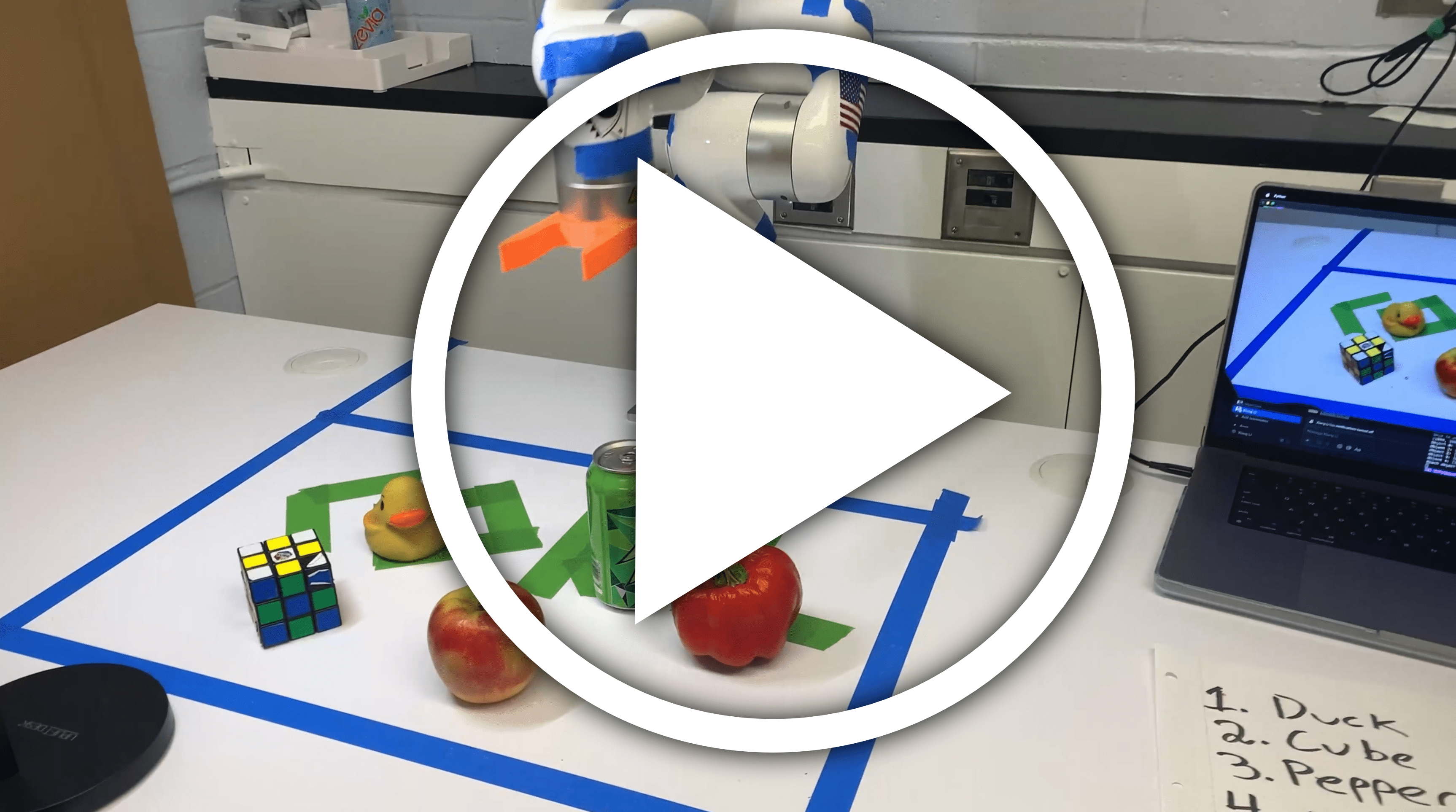}
    		}
    	\end{center}
    	\caption{We've created a video of the robot arm finding various objects, using the policy learned using TRITON. To watch, click the image or visit  \url{https://youtu.be/IwAs420hY6A}.
    	}
    	\label{fig:robot_five_robot}
    \end{figure}

    To visualize the outputs of TRITON and all of its baselines on this dataset, we have created an animation of those objects moving around so you can easily see the difference in temporal consistency between our algorithm and the baselines (See Figure~\ref{fig:robot_five_animation}). Figure~\ref{fig:robot_five_robot} and its corresponding video demonstrate the robot running in the real world.
    
\subsection{Deformable Object Experiment} 
\label{sec:american_flag}
    \paragraph{Summary} In this experiment, we train various algorithms to translate between synthetic UVL images and synthetic images of an American Flag blowing in the wind. This experiment demonstrates how TRITON handles deformable objects better than our baselines (MUNIT, CycleGAN and CUT) by comparing their outputs to a ground truth image. 
    \paragraph{Dataset} The AmericanFlag dataset comprises of 2000 training UVL scenes, and 80 (unpaired) photographs. For the testing data, we use 2000 UVL scenes and 2000 paired ground-truth photographs. 
    The UVL images and photographs are all generated in Blender, a 3d rendering program. We're using simulated photographs so we can easily get quantitative metrics by comparing the translated image with a ground truth. Since there is one flag and one background image, there are a total of 2 different labels: $L=2$. The height of the images are all 512 pixels. 
	\begin{figure}[H]
		\begin{center}
			\includegraphics[width=250pt]{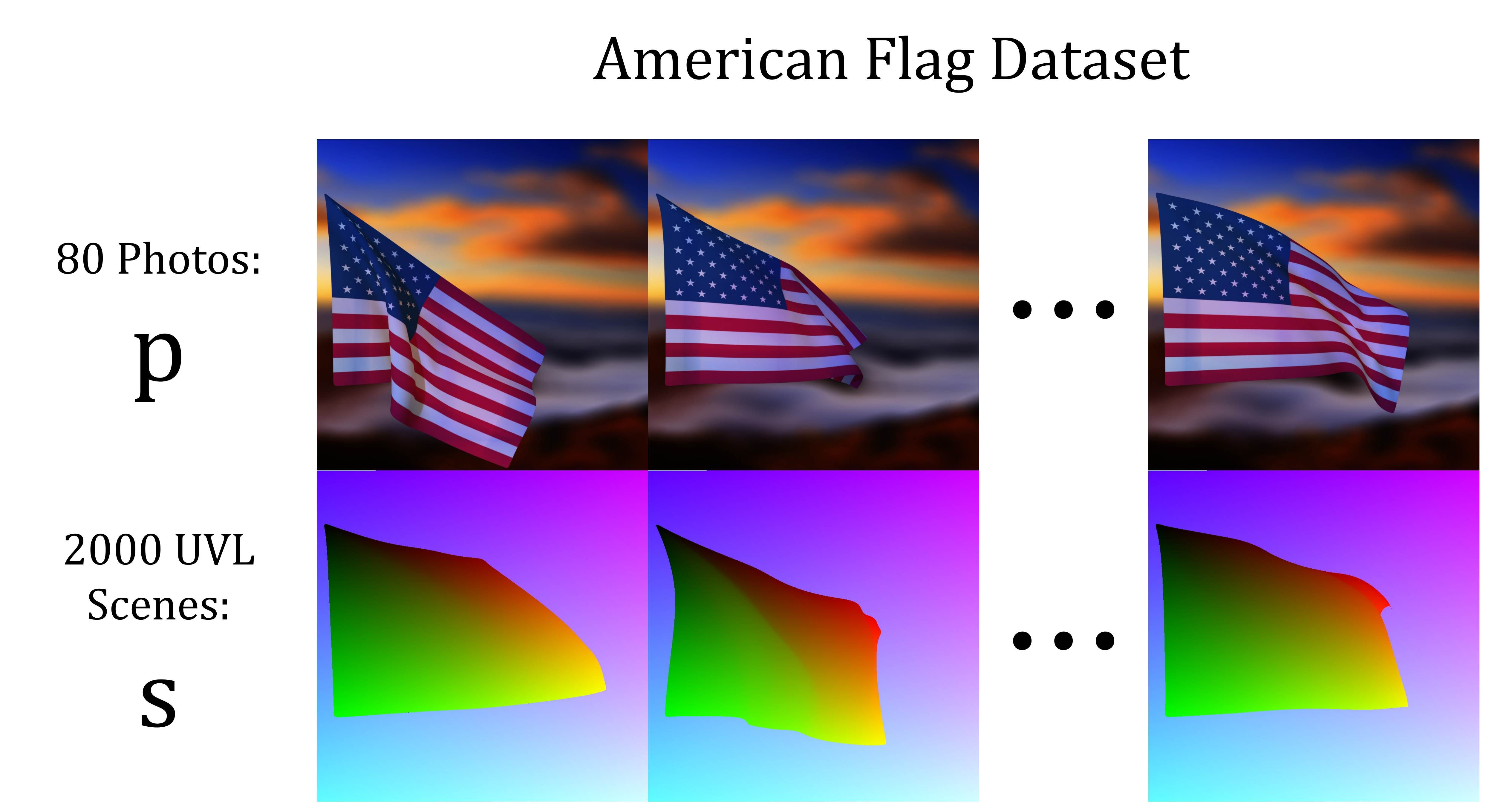}
		\end{center}
		\caption{
			This figure shows some random training samples in the AmericanFlag dataset.
		}
		\label{fig:americanFlagDataset}
	\end{figure}
	
    \paragraph{Results}
    We found that TRITON vastly outperforms our baselines according to both quantitative and qualitative results. In Table~\ref{tab:americanflag_results} we show that TRITON has much better quantitative results, and in~Figure~\ref{tab:americanflag_results} we show that TRITON preserves the stripes and stars better than our other baselines. For example, MUNIT may look realistic at first glance, but at some frames it doesn't include two of the thirteen red and white stripes. 
    
	\begin{table}[h]
	    \centering
	    \setlength{\tabcolsep}{2pt}
	    \caption{Quantitative results on the AmericanFlag dataset. We report the average perceptual image similarity (LPIPS), multiscale structural image similarity (MSSSIM) and $l2$-norm between the translated images and the ground truth images among all samples in the test set.}
	    \label{tab:americanflag_results}
        \begin{tabular}{lccc}
            \toprule
                                                  & L2 $(\times 10^{-2})\downarrow$ & LPIPS $(\times 10^{-1})\downarrow$ & MSSSIM $\uparrow$ \\ \midrule
            {MUNIT}           & 5.43                            & 3.778                              & .5901             \\
            {CycleGAN}        & 8.58                            & 4.833                              & .5600             \\
            {CUT}             & 8.72                            & 5.737                              & .4736             \\ 
            \midrule
            {\textbf{TRITON}} & \textbf{0.89}                   & \textbf{0.722}                     & \textbf{.9226}\\
            \bottomrule
        \end{tabular}
	\end{table}
    
	\begin{figure}[H]
		\begin{center}
		\href{https://youtu.be/HPk4PHLaut4}{
				\includegraphics[width=400pt]{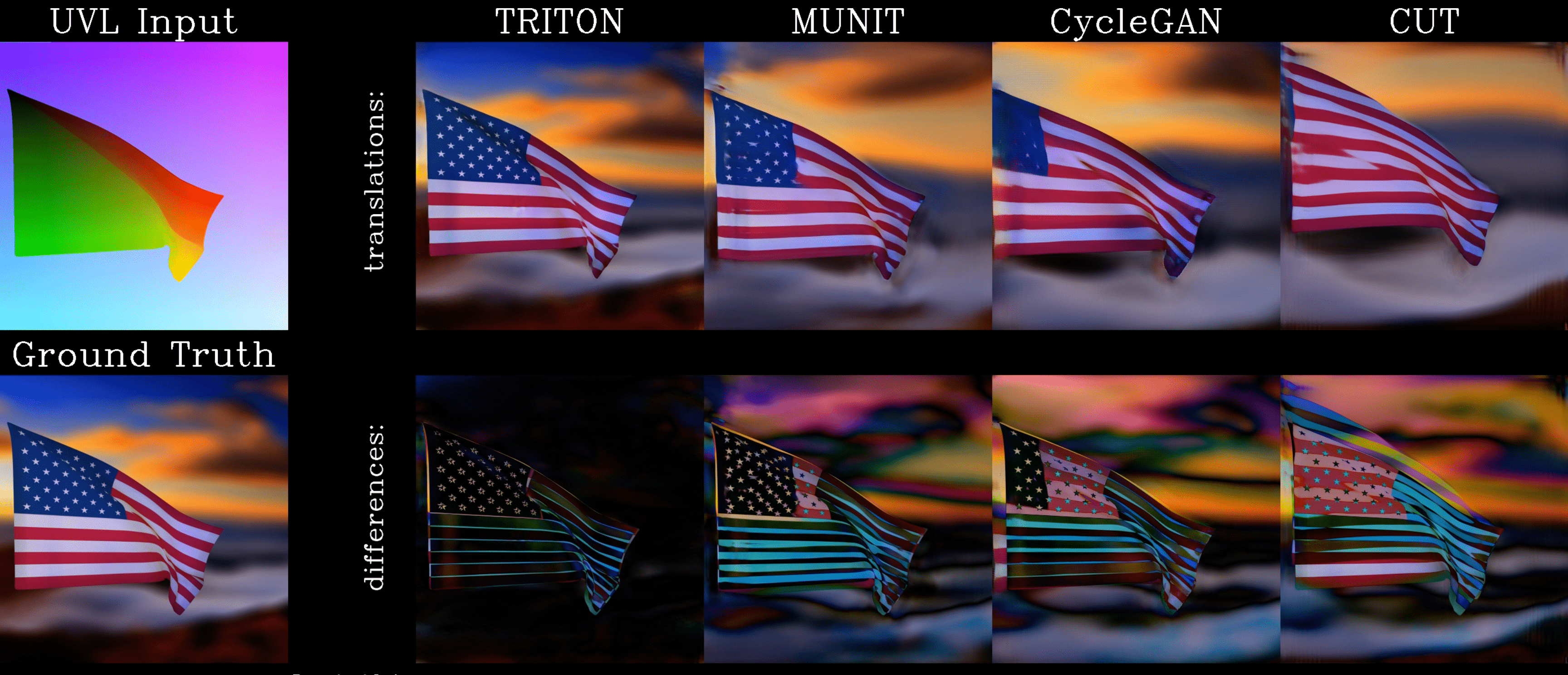}
				}
			\end{center}
		\caption{
			We have uploaded a video showing an animation of the AmericanFlag animation on all of our baselines, along with per-frame losses. Please click the above picture, or use the following link to view it: \url{https://youtu.be/HPk4PHLaut4}
		}
		\label{fig:robotarm_anim}
	\end{figure}

\subsection{Synthetic Object Experiment} 
\label{sec:threesynth}
\begin{figure}[H]
		\begin{center}
		\href{https://youtu.be/p6Mt0l8OPvw}{
				\includegraphics[width=400pt]{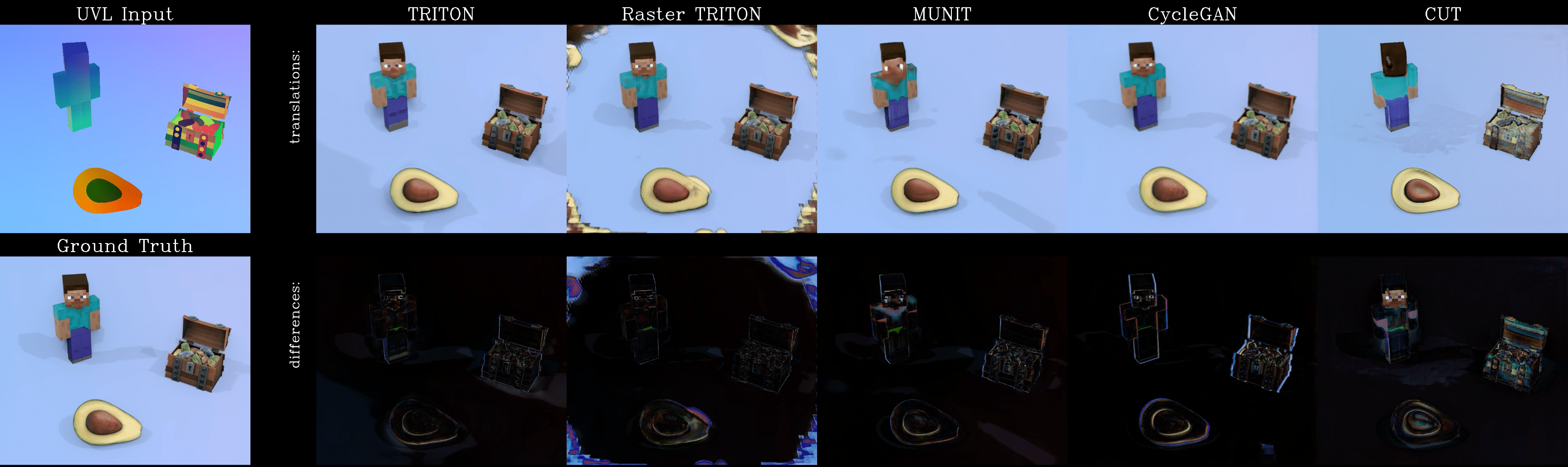}
				}
			\end{center}
		\caption{
			This figure shows an example of the translations on the ThreeSynth dataset. Click the image to view an animation of this analysis for three different datasets, or go to \url{https://youtu.be/p6Mt0l8OPvw}.
		}
		\label{fig:threesynth_anim}
	\end{figure}

    \paragraph{Summary} In this experiment, we train various algorithms to translate between synthetic UVL images and synthetic images of three rigid objects moving around a table. Because we're using a synthetic dataset, we can directly measure the accuracy of the translations across baselines. We show that TRITON outperforms the other algorithms.
    \paragraph{Dataset} The ThreeSynth training dataset comprises of 2000 training UVL scenes, and 500 (unpaired) photographs. For the testing data, we use 1795 UVL scenes and 1795 paired ground-truth photographs that were rendered in Blender. 
    As in the AmericanFlag dataset seen in Figure \ref{fig:americanFlagDataset}, we use simulated photographs so we can easily evaluate accuracy by comparing the translated image with a ground truth. Because the ThreeSynth dataset uses asymmetrical objects unlike the AlphabetCube-3 dataset, we do not need to match through many permutations to measure the results. Since there are three objects and one table, there are a total of 4 different labels: $L=4$. The height of the images are all 512 pixels. 
	\begin{figure}[H]
		\begin{center}
			\includegraphics[width=400pt]{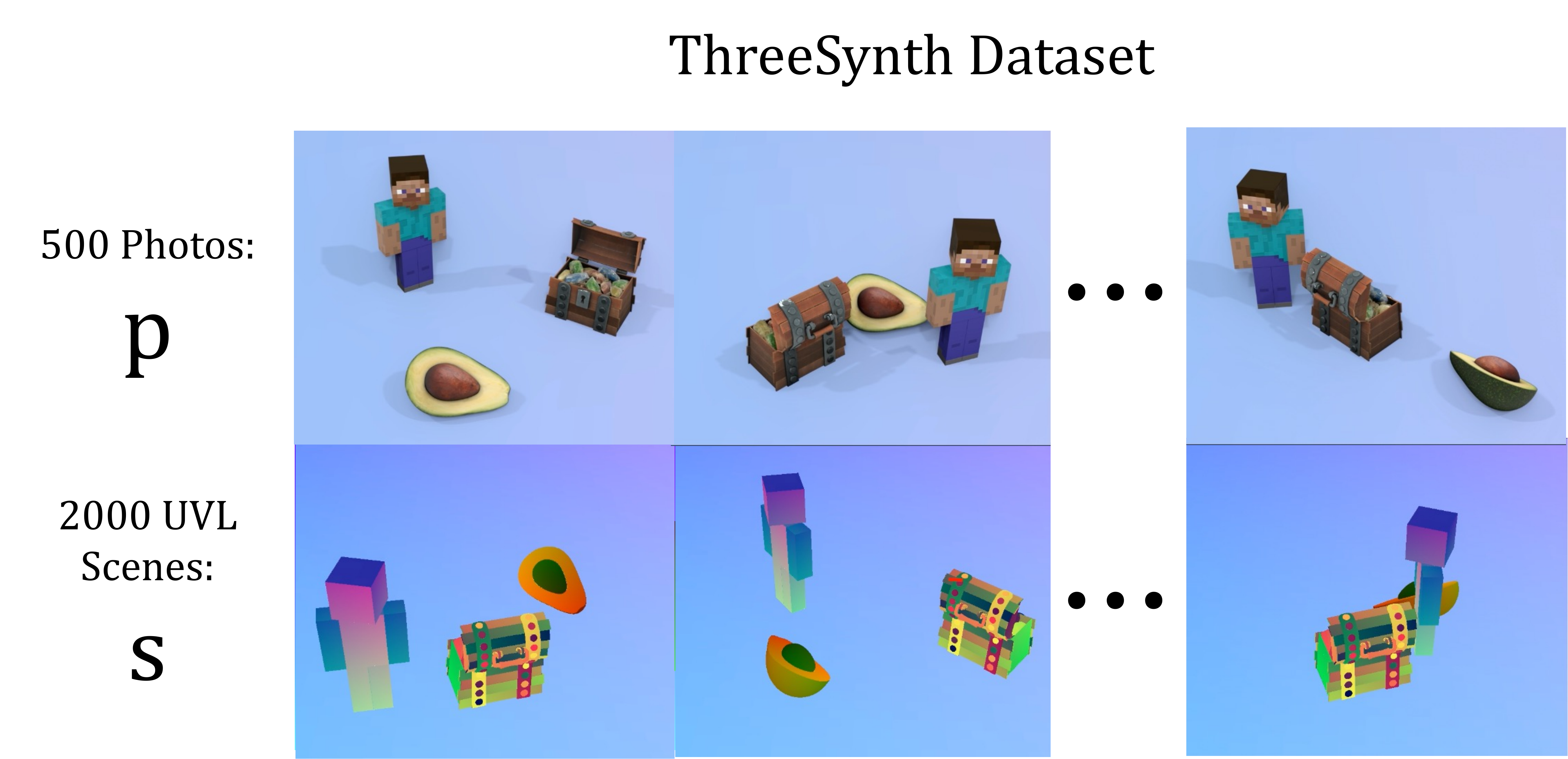}
		\end{center}
		\caption{
			This figure shows some random training samples in the ThreeSynth dataset. It includes three obejcts: A half-avocado, a treasure chest, and model of Steve from the game ``Minecraft''. 
		}
		\label{fig:threesynthdataset}
	\end{figure}
    \paragraph{Results}
    We found that TRITON outperforms our baselines according to both quantitative and qualitative results. In Table~\ref{tab:threesynthexperiments} we show that TRITON has better quantitative results, and in Figure~\ref{fig:threesynth_anim} we provide an animation of the objects moving. We observe that MUNIT distorts Steve's face.

   \begin{table}[h]
       \centering
       \setlength{\tabcolsep}{2pt}
       \caption{Quantitative results on the ThreeSynth dataset. We report the average perceptual image similarity (LPIPS), multiscale structural image similarity (MSSSIM) and $l2$-norm between the translated images and the ground truth images among all samples in the test set.}
       \label{tab:threesynthexperiments}
        \begin{tabular}{lccc}
        \toprule
            & L2 $(\times 10^{-3})\downarrow$ & LPIPS $(\times 10^{-1})\downarrow$ & MSSSIM $\uparrow$ \\ 
            \midrule
            {CUT}             & 4.07             & 0.8640            & .91419             \\
            {CycleGAN}        & 2.62             & 0.7979            & .93345             \\
            {MUNIT}           & 1.59             & 0.6754            & .95468             \\
            {Raster TRITON}   & 4.01             & 1.4049            & .95571             \\ 
            \midrule
            {\textbf{TRITON}} & \textbf{1.19}    & \textbf{0.6021}  & \textbf{.96221}       \\
        \bottomrule
        \end{tabular}
   \end{table}
	
\pagebreak
\section{Curiosities} 

In the curiousities section, we show what happens when you apply the wrong textures to the wrong objects, and show a way to isolate the shadows generated by TRITON. We've decided to include them in the appendix simply because they are interesting.
\subsection{Swapping Neural Textures}
What happens when you apply the wrong textures to the wrong objects? 

Below, we show what happens when you swap these textures \textit{after} training. Since the neural textures are evolved per-object, swapping labels during inference will usually not result in realistic translations. This is because the UV mapping for the neural textures will no longer match the geometries of the respective objects. Below we show an example of this using the RobotFive dataset.

\begin{figure}[h]
	\begin{center}
		\includegraphics[width=400pt]{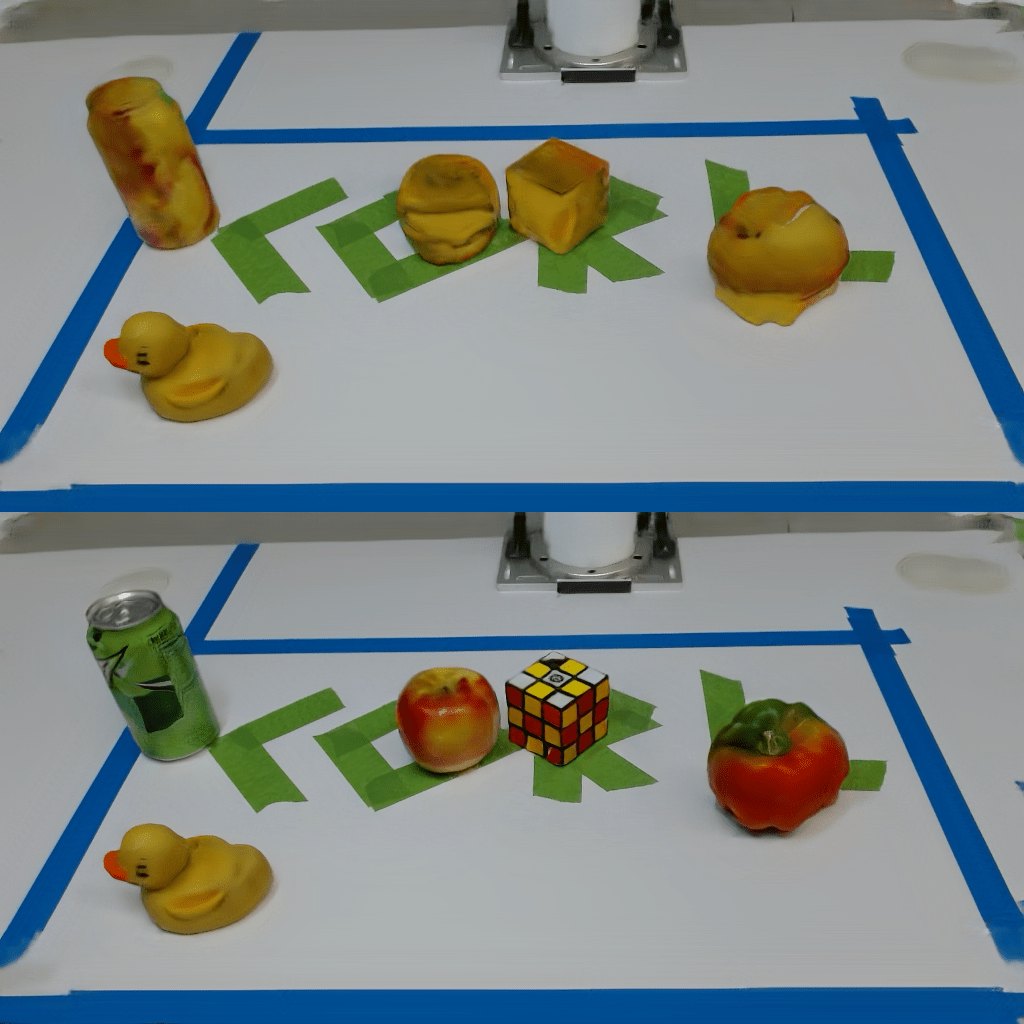}
	\end{center}
	\caption{
		On the bottom, we have a normal translation: it includes five objects (a soda can, an apple, a Rubik's cube, a pepper and a duck). On the top we've set the label of all objects (besides the table) to the duck label. It looks pretty terrible, because the UV mapping was never meant for that geometry. Please note that no rubber duckies were harmed in the making of this experiment.
	}
	\label{fig:texture_swap_to_duck}
	
\end{figure}

But what happens if we swap them \textit{during} training? We can get more interesting results if we swap the labels and then allow it to train for a while afterwards. Typically the objects will get stuck in a local minimum, and try to make the best of their new identities. This works because TRITON is fairly robust to geometric imperfections - it even manages to make the duck look like a Rubik's cube, and the apple to look like a very fat duck.

\begin{figure}[H]
	\begin{center}
			\href{https://youtu.be/ivfBslj4Gsg}{
		\includegraphics[width=400pt]{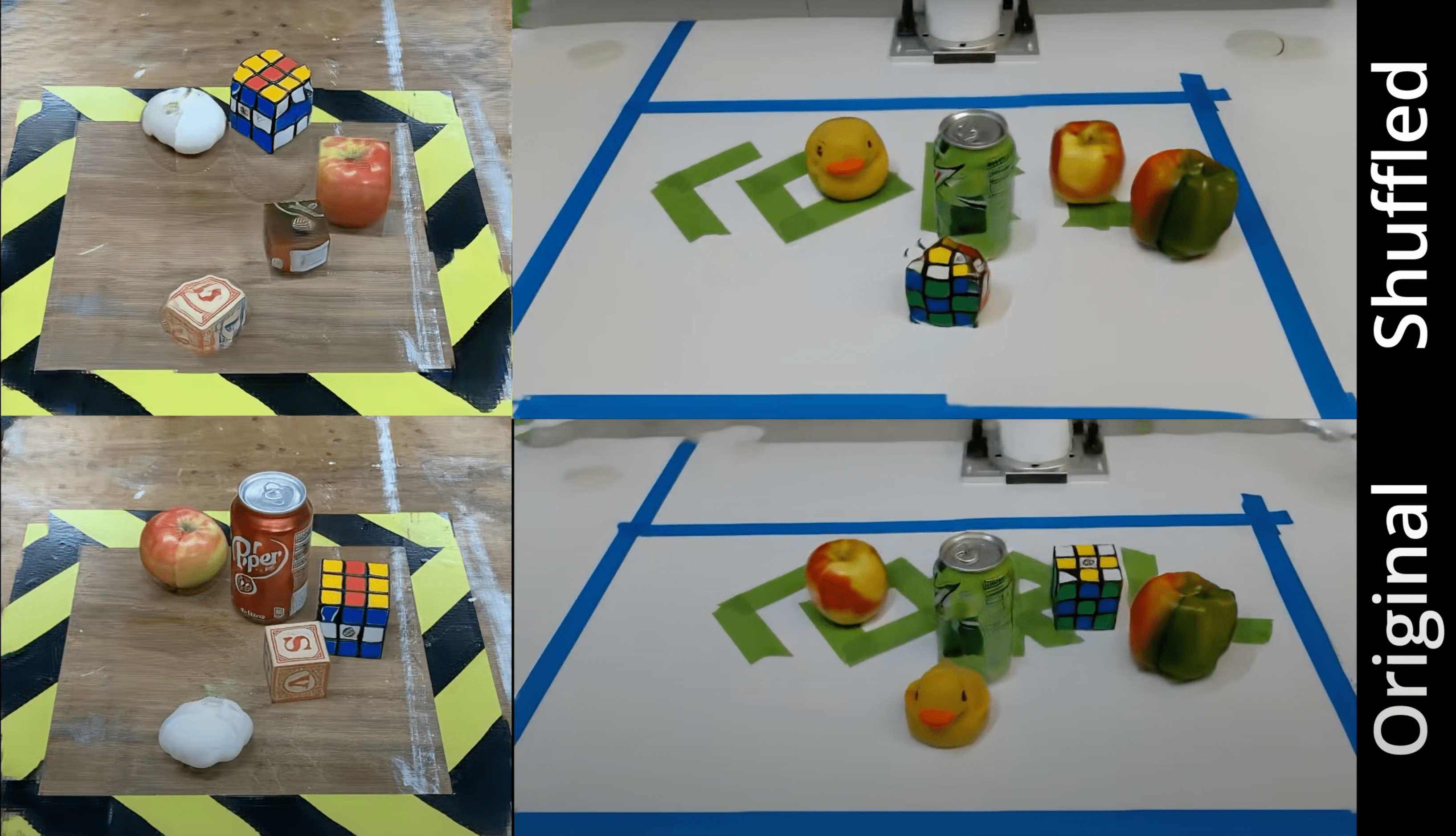}
		}
	\end{center}
	\caption{
		On the bottom two images, we have the original TRITON output for two different datasets. On the top, we've shuffled some of the labels and allowed it to train for an additional 50,000 iterations - letting it converge on a local minimum where the objects now have the wrong labels but still attempt to look good. On the left dataset, we shuffled the garlic, apple, soda, apple and alphabet cube labels. On the right dataset, we shuffled the duck, apple and rubiks cube labels. Click the image to view an animation, or go to \url{https://youtu.be/ivfBslj4Gsg}. Even the duck has managed to turn itself into an oddly convinving rubik's cube.	}
	\label{fig:label_shuffle_evolved.png}
\end{figure}

\subsection{Shadow Analysis}

TRITON can mimic shadows and shading effects, but they are not physically accurate. This is because, like the other baselines, TRITON is an image translation algorithm and does not have access to the latent geometry used to create a UVL scene image. Without the underlying geometry for a scene, it can't calculate the correct position for shadows. The fake shadows that TRITON generates are important for the output's realism though.

To isolate the shadows and shading artifacts, we randomly choose $100$ samples from the UVL dataset and translate them using TRITON. The objects in these samples have different orientations and locations so that the shadow is seen at different parts of the object in different samples.
Then we unproject the pixels of all the translated samples back onto a texture as seen in figure \ref{fig:unprojection_consistency_loss}, and the average of all the unprojected textures are denoted as the mean unprojected texture $\mu(\unprojection)$.
By averaging over $100$ samples, we can get rid of the random noisy shadow in a single image and recover the original texture agreed by majority of the samples without shading artifacts. 
Then the mean unprojected texture $\mu(\unprojection)$ is projected back and this image without shadow is called ``Reprojection'' (See the bottom right figure in Figure~\ref{fig:label_shuffle_evolved.png}).
Finally the shadow of an individual sample can be extracted by subtracting ``Reprojection'' from the sample (Top right in Figure~\ref{fig:label_shuffle_evolved.png}).

\begin{figure}[H]
	\begin{center}
			\href{https://youtu.be/ZscnUbKOxdg}{
		\includegraphics[width=300pt]{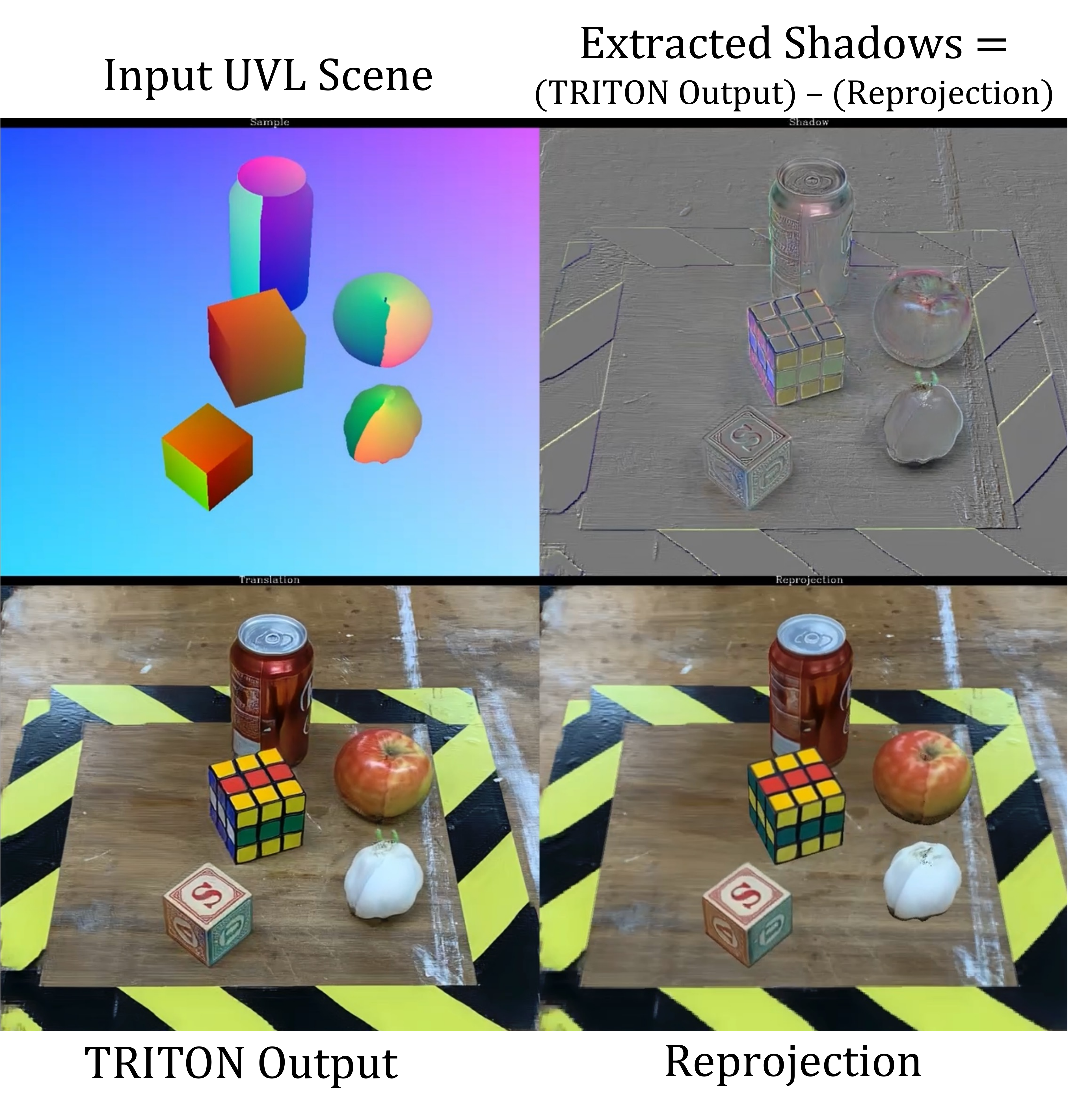}
		}
	\end{center}
	\caption{
		In this figure, we show the extracted shadows for a given scene. By looking at the 'Shadows' corner of the image, you can see that there are dark regions under the can and the apple - indicating that those shadow values are below the average surface value for that dataset's TRITON output. Note that it also shows other lighting effects beyond shadows, such as the shiny parts of the soda can. Click the image to view an animation of this analysis for three different datasets, or go to \url{https://youtu.be/ZscnUbKOxdg}.
	}
\end{figure}